\documentclass{article}
\usepackage{natbib}
\usepackage{float}
\usepackage{microtype}
\usepackage{graphicx}
\usepackage{subfigure}
\usepackage{booktabs} % for professional tables
\usepackage{bbm}
\usepackage{hyperref}

\usepackage{amsmath}
\usepackage{amssymb}
\usepackage{mathtools}
\usepackage{amsthm}

\usepackage[capitalize,noabbrev]{cleveref}

\theoremstyle{plain}
\newtheorem{theorem}{Theorem}[section]
\newtheorem{proposition}[theorem]{Proposition}
\newtheorem{lemma}[theorem]{Lemma}

\theoremstyle{definition}
\newtheorem{definition}[theorem]{Definition}
\newtheorem{assumption}[theorem]{Assumption}
\theoremstyle{remark}

\newcommand{\sgn}{\operatorname{sgn}}

\usepackage[textsize=tiny]{todonotes}
\usepackage[letterpaper,margin=1in]{geometry}  % If using US letter size

\title{Provable Benefits of Unsupervised Pre-training and Transfer Learning via Single-Index Models}

\author{
  Taj Jones-McCormick\thanks{Department of Statistics and Actuarial Science, University of Waterloo, Canada. Equal contribution.}, 
  Aukosh Jagannath\thanks{Department of Statistics and Actuarial Science, University of Waterloo, Canada; Cheriton School of Computer Science, University of Waterloo, Canada. Equal contribution.}, $\And$
  Subhabrata Sen\thanks{Department of Statistics, Harvard University, United States of America. Equal contribution.}
}

\begin{document}
\maketitle

\vskip 0.3in

\begin{abstract}

    Unsupervised pre-training and transfer learning are commonly used techniques to initialize training algorithms for neural networks, particularly in settings with limited labeled data. In this paper, we study the effects of unsupervised pre-training and transfer learning on the sample complexity of high-dimensional supervised learning. Specifically, we consider the problem of training a single-layer neural network via online stochastic gradient descent. We establish that pre-training and transfer learning (under concept shift) reduce sample complexity by polynomial factors (in the dimension) under very general assumptions. We also uncover some surprising settings where pre-training grants exponential improvement over random initialization in terms of sample complexity.  
\end{abstract}

\section{Introduction}
The canonical pipeline of modern supervised learning is as follows: given supervised data, (i) choose an appropriate model/estimator (usually specified by a deep neural network), (ii) choose a loss function and set up a suitable empirical risk minimization problem, and (iii) minimize this (possibly non-convex) empirical risk  using stochastic gradient descent (SGD). Extensions of this ``basic" approach have been successfully deployed to train state-of-the-art models in diverse domains. As deep learning models become larger and more complex, one has to wrestle with the issue of model weight initialization during training. Without additional information, one usually resorts to random initialization. However, access to additional data opens the door to other avenues for initializing model weights.

A prominent setting with additional data is semi-supervised learning, where one might have an abundance of unlabeled data. Unsupervised pre-training has emerged as a popular strategy in this context \cite{devlin2019bert,brown2020language}. The core idea behind pre-training is  to train a model (on the unlabeled data) on a task that is related or, even better, a necessary precursor to the supervised task of interest. Initializing from a pre-trained model, the hope is that the model will have already learned useful features from the unlabeled data and hence solve the supervised task with reduced sample complexity. The model trained in this way can then be used to initialize a model for the supervised task in various ways e.g., by removing the final layer of the network and replacing it with an output head relevant to the labeled task, such as a classification head. 

Another prominent setup with additional data is transfer learning \cite{pan2009survey}, where one has access to samples from related supervised tasks. In this case, a natural idea is to initialize the model weights on the target dataset using the model trained on the upstream task.

The precise scheme for unsupervised pre-training or transfer learning can vary significantly across applications. For example, BERT models employ self-supervised representation learning by predicting ``masked" tokens based on the observed ones \cite{devlin2019bert}. Similar self-supervised pre-training algorithms form core components in the training of modern language models such as GPT \cite{radford2019language, brown2020language, achiam2023gpt} and have attracted widespread attention recently. Many other forms of both pre-training and transfer leaning are applied in countless works spanning many different fields within machine learning \cite{wang2016learning, he2017mask, devlin2019bert, schneider2019wav2vec}. Their popularity underscores the importance of understanding the effects of these different methods of weight initialization on supervised tasks.

The main goal of this work is to build towards a theoretical understanding of the benefits of pre-training, particularly in terms of the effect on sample complexity for solving supervised learning tasks in high dimension that involve optimizing non-convex losses. In general, characterizing the performance of neural networks on supervised learning tasks is a challenging problem. Recent works  focus on specific classes of problems such as single-index models learned with single-layer networks \cite{arous2021online} or two-layer networks \cite{lee2024neural}, and characterize the number of samples required for recovery of the latent signal. We study the effects of distinct initializations on the sample complexity for a closely related class of problems. Specifically, we show provable benefits of pre-training and transfer learning  in terms of reducing sample complexity for single-layer networks. We also highlight surprising complexities and powerful benefits of pre-training---we discover simple scenarios under which one cannot hope to succeed with random initialization, but the problem can be solved easily with suitable pre-training. Finally, we demonstrate our findings empirically in finite dimensional settings with simulations.

\section{Related Work}

%Firstly we will discuss recent theoretical works attempting to further our understanding on the benefits of pre-training or transfer learning, and how these works differ from ours. Next, we will discuss recent works containing results on convergence and required sample complexity for solving supervised tasks with SGD. We will conclude with a brief discussion of work on single index models.
We summarize some related works in this section, and compare these prior works with the contributions in this article.

\subsection{Pre-training and Transfer Learning Theory}
There has been significant recent progress in understanding the benefits of distinct unsupervised pre-training methods. In \citet{lee2021predicting}, the authors provide rigorous evidence of the benefits of self-supervised pre-training (SSL). They explain the benefits of SSL via specific conditional independence relations between two sets of observed features, given the response. In a related direction, \citet{arora2019theoretical,tosh2021contrastive,tosh2021contrastive2} examine the benefits of contrastive pretraining, while \citet{zhai2023understanding} examines the effects of augmentation-based self-supervised representation learning. In \citet{wei2021pretrained}, the authors explore the benefits pre-trained language models, while  \citet{zhang2021inductive}, explores the inductive bias of masked language modeling by connecting to the statistical literature on learning graphical models. Finally, we highlight the work \cite{azar2024semi}, which  exhibits provable computational benefits of semi-supervised learning under the low-degree likelihood hardness conjecture.

The paucity of high-quality labeled data has directly motivated inquiries into the properties of transfer learning across diverse application domains. The recent literature focuses on several distinct notions of transfer learning (e.g., covariate shift \cite{heckman1979sample,huang2006correcting,shimodaira2000improving}, model shift \cite{wang2015generalization,wang2014active}, target shift \cite{maity2022minimax}, conditional shift  \cite{quinonero2022dataset,storkey2008training} etc) and develops distinct rigorous methods to ensure successful knowledge transfer in these settings (see \cite{shimodaira2000improving, ganin2016domain, long2017deep, wu2019domain, sagawa2019distributionally, gerace2022probing} and the references therein for an incomplete list). From a learning theoretic perspective, recent works study the generalization performance as a function of the discrepancy between the source and the target domains \cite{albuquerque2019generalizing,ben2010theory,david2010impossibility,hanneke2019value,tachet2020domain,zhao2019learning}.

In \cite{damian2022neural}, the authors study the benefits of transfer learning in the setting of single/multi-index models. They keep the representation fixed across the source and target, and vary the link function across the two tasks. In contrast, we keep the link function constant (and assume that the link is known), and study settings with distinct (but correlated) representations in the source and target tasks.

\subsection{Understanding Sample Complexity for single-index models}

% The goal of pre-training is to reduce training time and the number of samples required to estimate the function of interest. To understand this benefit, it is of fundamental importance to understand the sample complexity required to solve problems with stochastic gradient descent (SGD) in high dimensions.

 Single-index models have emerged as popular toy-models for understanding the sample complexity of training of neural networks. This is due to the fact that they are both high-dimensional and non-convex. From a statistical perspective there has been work on the fundamental thresholds of inference in these problems \cite{barbier2019optimal,maillard2020phase} and its landscape geometry \cite{sun2018geometric,maillard2019landscape,dudeja2018learning}. From the perspective of sample complexity, a substantial amount of deep work in this direction focused on the sample complexity  of spectral methods or related algorithms, particularly in relation to the Phase Retrieval problem, \cite{candes2015phase,barbier2019optimal, lu2020phase}.  
% It is important to note that, while we model the pre-training step via PCA, the perspective and assumptions are fundamentally different. Those works focus on the supervised setting, where one performs the spectral method on matrices that depend on both labels and features (i.e., during training) whereas in our work we seek to model the pre-training phase using unlabeled data via optimizing a quadratic loss (equiv. PCA or online PCA) only given the features.

More recently there have been tight analyses of the sample complexity for online stochastic gradient descent from random initialization.
In particular, it was shown in \cite{arous2021online} that the sample complexity in the online setting is characterized by the Information Exponent. Since then there has been a tremendous body of work around complexity exponents, such as the Information Exponent, Leap Exponent \cite{abbe2023sgd}, or Generative Exponent \cite{damian2024computational}. In particular, these exponents have enabled studies which contrast the performance of various learning paradigms such as Correlational Statistical Query (CSQ) versus Statistical Query (SQ) bounds \cite{damian2024computational}, feature learning versus kernel methods \cite{ba2024learning}, better choices of loss function \cite{damian2024smoothing}, and the importance of data reuse \cite{dandi2024benefits,lee2024neural, arnaboldi2024repetita}.  We note here that there has been quite a lot of recent important work on the case of multi-index models which we do not explore here, see, e.g., \cite{abbe2023sgd,bietti2023learning,ren2024learning} for a small selection of this rich literature.  

To our knowledge, most of this work has focused largely on the setting of isotropic Gaussian features (though note \citet{zweig2024single, pesce2023gaussian} for work on universality). However, given that pre-training only has access to the features, one requires that the features have some correlation with the underlying spike. Inspired by the recent works of \cite{mousavi2023gradient, ba2024learning}, we model this via a spiked covariance model. There are some additional recent works cite{dandi2024random, cui2024asymptotics, ba2022high} which analyze the high dimensional asymptotics of neural network training after a single gradient step and also consider spiked and isotropic index models.

% in high-dimensional settings. From the perspective of the sample complexity, there has benen a tremendous amount}

% There are various efforts towards understanding this problem, particularly when considering single index models. The information exponent \cite{arous2021online} is a quantity which governs the sample complexity for estimating single index models with single layer networks using online SGD. These results are expanded upon by \cite{damian2024smoothing}. When estimating single index models with two layer neural networks, the complexity is governed by the generative exponent \cite{lee2024neural}. In \citet{abbe2023sgd}, high dimensional models with a low dimensional latent structure are also considered and they prove results regarding the complexity of estimation with 2-layer networks based on the leap exponent of the target function. 

% \subsection{Single Index Models}

% The models we consider in our work are single index models. These models have been studied extensively in many works, some of which include \citet{mondelli2018fundamental}, \citet{sun2018geometric}, \citet{barbier2019optimal}, \citet{maillard2019landscape}, \citet{lu2020phase}. Many works consider these models with isotropic Gaussian features (\citet{arous2021online}, \citet{lee2024neural}), although recent works have also considered Gaussian features with spiked covariance structure \cite{ba2024learning}, both of which we consider in this paper.

\section{Pre-Training}

\subsection{Problem Set Up and Notation}

%We now introduce the statistical model we will study under pre-training. 
We consider a single layer supervised network with specified activation function $f$. We consider Gaussian features with spiked covariance, where the spike is correlated with the parameter vector of interest. %This is a single index model.

Let the labeled data be $(y_i, a_i)_{i=1}^N$, with each $(y_i, a_i)$ independent and identically distributed. We have the following relationship between $a$ and $y$: $y_i = f(a_i \cdot v_0) + \epsilon_i$, for some $\epsilon_i$ independent of $a_i$, with mean 0 and finite fifth moment. The parameter vector we wish to estimate is $v_0$ and $f$ is a known activation function. Throughout we assume that $f$ is twice differentiable almost everywhere with $f,f',f''$ of at most polynomial growth. We would like the model we consider to capture the essence of pre-training. To perform pre-training, one should have access to additional unlabeled data. We thus assume access to some unlabeled $(a_{i}')_{i=1}^{N'}$ with $a_{i}' \stackrel{D}{=} a_{i}$. In order for pre-training to be useful, there is an implicit assumption that the unlabeled data contains some information in its structure that is related to the supervised task. We thus let $a_i \sim N(0, I_d +\lambda vv^T)$, with $v \cdot v_0 = \eta_1 \in [0,1], \lambda > 0$ ($\eta_1, \lambda$ are dimension independent) and $v,v_0 \in \mathbb{S}^{d-1}$ (the unit sphere in $\mathbb R^d$). Thus, the features are Gaussian with spiked covariance, where the spike vector has some correlation with the unknown parameter vector of interest $v_0$. In this way, our model captures the significance of pre-training by allowing the unlabeled feature data to contain information hidden in its covariance structure that is directly correlated with the solution of the supervised learning problem. The value of $\eta_1$ measures the strength of this correlation. We define $\eta_2 = \sqrt{1 - \eta_1^2}$.

Our goal is to estimate the unknown vector $v_0$ with parameter vector $X \in \mathbb{S}^{d-1}$, by using SGD on the following loss function: $\mathcal{L}(X,y) = [f(X \cdot a) - y]^2$. We use spherical gradient descent with step-size $\delta/d$ to optimize parameters $X$, given by the following stochastic updates: $$X_{t+1} = \frac{X_t - \frac{\delta}{d}\nabla \mathcal{L}(X_t, y)}{\|X_t - \frac{\delta}{d}\nabla \mathcal{L}(X_t, y)\|_2}$$ with initialization $X_0$, where $\nabla$ denotes the spherical gradient with respect to the parameters $X$.

We want to understand the benefit of pre-training and so we consider two methods of initializations, random and with pre-training. For random initializations we let $X_0 \sim \mathrm{Uniform}(\mathbb{S}^{d-1})$. To model pre-training, we use Principal Component Analysis (PCA) on the unlabeled data $(a_{i}')_{i=1}^{N'}$, to obtain an estimate $\hat{v}$ of the spike direction $v$. We then use this to initialize SGD for our supervised task, that is we let $X_0 = \hat{v}$. PCA is arguably the ideal starting point for a rigorous investigation into unsupervised pre-training. From a statistical perspective, PCA is the simplest dimension-reduction algorithm; further, it's properties are well-understood in high-dimensions \cite{bai2010spectral}. More importantly, it has been shown that more advanced representation learning algorithms, such as reconstruction autoencoders, also implement PCA in certain regimes \cite{bourlard1988auto, baldi1989neural,nguyen2021analysis}. In this light, we will restrict ourselves to PCA-based pre-training in this paper.    We note that this approach is distinct from the spectral methods introduced for single-index models and, in particular, phase retrieval. There, the methods are supervised in that they use both knowledge of the label and features, whereas in unsupervised pre-training one only has access to the features. 

With these two methods of initialization, our goal is to contrast their respective sample complexity requirements for solving the supervised learning task (recovering the unknown vector $v_0$). We are in particular interested in the high dimensional regime. We consider spherical SGD (hence forth referred to as simply SGD) with the total number of steps (and samples of $(y_i, a_i)$) given by $N = \alpha_d d$. Thus the number of samples we observe is a function of the dimension, and we are then interested in analyzing the high dimensional limit $d \to \infty$.

\subsection{Main Results}
\label{main}

 In this section we state our main results. Firstly we state a few definitions and assumptions. We defer the proofs of all results to Appendix \ref{appendix}. Throughout we will often refer to the `population loss': 
 \begin{align*}
 \Phi(X) &= \mathbb{E}[f(X \cdot a) - y]^2 \\
 &= \mathbb{E}[f(X \cdot a) - f(v_0 \cdot a)]^2 + \mathbb{E}\epsilon^2
 \end{align*}
 We note that there are two important directions of interest in this problem, namely $v_0$ and the residual direction of the spike vector, after subtracting off the projection onto $v_0$, that is $\frac{1}{\eta_2}(v - \eta_1 v_0)$. Without loss of generality, we let the first two basis vectors be written as $e_1 = v_0$ and $e_2 = \frac{1}{\eta_2}(v - \eta_1 v_0)$, so that $v = \eta_1 e_1 + \eta_2 e_2$. We can rewrite the population loss which is a function of $X$, solely through the correlation of $X$ with each of these directions. Let $m_1(X) = X \cdot e_1 = x_1$ and $m_2(X) = X \cdot e_2 = x_2$, then 
\begin{align*}
     &\Phi(X)  = \mathbb{E}[f(a \cdot X) - f(a \cdot e_1)]^2 + \mathbb{E}\epsilon^2 \\ 
     & =  \mathbb{E}[f(a_1 x_1 + a_2 x_2 + \sqrt{1 - x_1^2 - x_2^2}g) - f(a_1)]^2 + \mathbb{E}\epsilon^2 \\
    & = \phi(x_1, x_2) 
\end{align*}

where $[a_{1},a_{2},g]$ is jointly Gaussian with mean 0 and $g \perp a_{1},a_{2}$ and $Eg^2 = 1$. The covariance of $a_{1},a_{2}$ is given by: $$\begin{pmatrix}
1 + \lambda \eta_{1}^2 & \lambda \eta_{1} \eta_{2}    \\
\lambda \eta_{1} \eta_{2} &  1 + \lambda \eta_{2}^2   \\
\end{pmatrix}$$

Throughout we will use the term `population flow' which is simply the discretized gradient flow on the population loss $\Phi$. 
% as obtained by the `ODE method' \cite{ljung1977analysis}. We define the population flow as the following process: $$\bar{X_{t}} = \frac{\bar{X}_{t-1} - \frac{\delta}{d}\nabla \Phi(\bar{X}_{t-1})}{\| \bar{X}_{t-1} - \frac{\delta}{d}\nabla \Phi(\bar{X}_{t-1})\|_{2}}, \bar{X}_{0} = X_{0}$$ 
%In some cases we consider the process to be a 2-dimensional system in the two correlation variables of interest $m_1(X_t)$ and $m_2(X_t)$, which is sufficient to describe the dynamics of $\phi$ in $X$. 
%We do not specify this when the context is clear. We also make use the notation, $X_n \gg Y_n$ when $Y_n/X_n \to 0$.

\begin{definition} \label{def:effective}
    A sequence of initializations $(X_0^{(d)})_{d \geq 1}$ is \textbf{Effective} for SGD with $N = \alpha_d  d$ steps of stepsize $\delta/d$ if
    $m_1(X_N^{(d)})\to 1$ in probability as $d \to \infty$.
\end{definition}

A sequence of initializations is considered Effective if SGD with some number of steps and stepsize, initialized from the given sequence, recovers the true solution in the high dimensional limit. To that end, we say that a sequence of initializations is \textbf{Ineffective} if it is not effective. We have defined Effective initializations for SGD based on the convergence of the SGD process, and we can also consider these definitions for initializations of population flow, defined by the convergence of the population flow process in place of SGD.

\begin{assumption}\label{assumption:Assumption B}
    We say that \textbf{Assumption \ref{assumption:Assumption B}} holds with point $m^* = (m_1^*, m_2^*) \in B_2(0,1)$ if there exists a point $m^*$, such that:
    \begin{align*}
    &\nabla \Phi(X)\cdot e_1 > 0 \\
    &\sgn(m_2(X))\nabla \Phi(X)\cdot e_2  < 0
\end{align*}
for all $X$ such that $(m_1(X), m_2(X)) \in B_2(0,1) : m_1(X) \geq m_{1}^*,  \left|m_2(X)\right| < m_{2}^*$. Where $B_2(0,1)$ is the 2-d unit ball.

\end{assumption}

We note that the ability of $\Phi$ to meet Assumption \ref{assumption:Assumption B} depends entirely on the choice of activation function $f$. It is clear that population flow, initialized within the rectangle defined by the points $m^*$ and $(1,0)$ will recover the correct solution (see Figure \ref{fig:simulations} for an example), hence (emphasizing that $m^*$ is independent of the dimension) this assumption provides a simple way to verify that a sequence of initializations is Effective for population flow. Now we state our main result regarding initializing with pre-training.

\begin{theorem}
\label{thm:theorem1}
Suppose that Assumption \ref{assumption:Assumption B} holds with point $m^*$. Further $\eta_{1} \geq m^*_{1}$ and $\left|\eta_{2} \right| \leq m^*_{2}$. Then for spherical SGD on the given loss with $N = \alpha  d$ steps where $\alpha = \omega(1)$, $\alpha \delta^2 = o(1)$, we have that the sequence of initializations $X_0^{(d)} = \hat{v}_d$, the PCA estimators of $v_d$ obtained with $N' = \alpha'd$ unlabeled samples where $\alpha' = \omega(1)$, are Effective.
\end{theorem}

The theorem above states sufficient conditions on $\Phi$ and the correlation between $v$ and $v_0$ such that with pre-training, we are able to recover the true parameter vector with high probability in large enough dimensions. Further, we see that $\alpha = \omega(1)$, and hence our recovery with $N=\alpha d$ steps is just beyond linearly many steps in the dimension.

For our next two results we will work with activations $f$ which satisfy
\begin{equation}\label{eq:mom-cond}
    \mathbb{E}f''(g) = \mathbb{E}f'(g) = 0,\;\; \mathbb{E}\frac{\partial ^2}{\partial g^2}f(g)^2 > 0
\end{equation}
for $g \sim \mathcal{N}(0,1)$. Recalling the notion of information exponent from \citet{arous2021online} which is restated in the following section, (\ref{eq:mom-cond}) is equivalent to requiring the information exponent 3 or greater for $f$ and the information exponent less than or equal to 2 for $f^2$. We now consider our second main result which considers recovery from random initializations. 

\begin{theorem}\label{thm:theorem2}
Suppose that $f$ satisfies \eqref{eq:mom-cond}.
 Then for spherical SGD with $N = \alpha  d$ steps where $\alpha \ll d$, $\alpha \delta^2 = O(1)$, for the sequence of initializations $X_0^{(d)} \sim \mathrm{Uniform}(\mathbb{S}^{d-1})$ we have that $ m_1(X_N) \to 0$
in probability, as $d \to \infty$.
\end{theorem}

Our second main result states that under appropriate moment conditions on $f$, we have that in order to recover the unknown parameter vector $v_0$, we require at least $N = \Omega(d^2)$ samples. We emphasize that this result does not inform us of when we can recover the true parameter vector, only sufficient conditions for showing that we cannot recover with less than quadratic samples in the dimension. 

We now state our third main theorem which is the most surprising result and brings to light the complexity of the single layer supervised network with Gaussian features and spiked covariance. 
Let $H_r^d=\{X\in\mathbb S^{d-1}: |m_1(X)|<r\}$.
\begin{theorem}
\label{thm:theorem3}
Suppose that $f$ satisfies \eqref{eq:mom-cond}
for $g \sim \mathcal{N}(0,1)$. When $\eta_1 = 1$, for spherical SGD with $N = \alpha  d$ steps where $\alpha = \omega(1)$, $\alpha \delta^2 \ll d^{1/3}$ , we have that there exists some dimension independent value $r > 0$ such that for all sequences of initializations $X_0^{(d)}\in H_r^{(d)}$, then we have that:
$m_1(X_N) \to 0$ in probability as $d \to \infty$. Further we have that for all $\epsilon > 0$:
\begin{equation*}
    \mathbb{P}(\sup_{t \leq N}\left|m_1(X_t)\right| > r + \epsilon) \to 0 
\end{equation*}
in probability as $d \to \infty$.

\end{theorem}

This result demonstrates the surprising fact that in the simple scenario where the spike is perfectly aligned with the unknown parameter vector, recovery is not possible from random initializations with any amount of data, given appropriate stepsize. Even more surprising, not only is recovery not possible from random initialization, but even initializing with some fixed correlation, can result in not only a failure to recover but further a loss of the initial correlation. We also have that the maximum correlation attained over the course of SGD is contained in a ball around 0 with radius slightly larger than the ball containing the initializations. Taking into account Theorem \ref{thm:theorem1}, we see that there exists problems such that pre-training can allow us to solve the problem in linear time, whereas the problem is unsolvable from random initialization in the given scaling regime, regardless of the amount of labeled data.

It is important here to note that a similar negative result was observed by \citet{mousavi2023gradient} for population gradient flow when fitting a two-layer network with ReLU activation. There the authors propose to correct for this via preconditioning the gradient.  By contrast, here we use this to illustrate the power of pre-training.

\subsection{Discussion}

Together, the first two theorems above tell us that for certain activation functions $f$ such that both the assumptions of Theorem \ref{thm:theorem1} and the assumptions of Theorem \ref{thm:theorem2} are met, we establish a significant separation of the required samples for recovering the unknown parameter vector. With pre-training, we can achieve convergence with $N = \alpha d$ whenever $\alpha = \omega(1)$. That is, we can recover with less than log-linear samples in the dimension. For random initializations, we require $N = \Omega(d^2)$ at minimum. This gives us a separation of $d^\zeta$ for all $\zeta < 1$. Theorem \ref{thm:theorem3} not only highlights the complex scenarios that can arise by introduction of a spike vector but also serves as a demonstration of the powerful effects of pre-training. In the scenario where the spike vector is equal to the parameter vector, pre-training alone is sufficient for solving the problem (of course this information would not be available to any practitioner) and without pre-training, no amount of data is enough to solve the problem from random initialization under the given regime. However, provided enough correlation between the spike vector and unknown parameter vector, the problem is solvable via pre-training with just over linearly many samples.

We point out that Theorems \ref{thm:theorem1} and \ref{thm:theorem2} have established a lower bound on the benefit of pre-training. As we see from Theorem \ref{thm:theorem3}, there are scenarios which deviate significantly from the lower bound provided here. Past works \cite{arous2021online} have shown that when the features are isotropic Gaussian, the sample complexity is governed by a quantity called the information exponent, which is essentially the order of the first non-zero term in the Taylor expansion of the population loss. In the case of single-index models the information exponent can be written in terms of the Hermite coefficients of the activation function $f$, which can be similarly expressed as moment conditions on $f$. In light of the results in the isotropic Gaussian case, Theorems \ref{thm:theorem1} and \ref{thm:theorem2} may not seem that surprising. We emphasize here that the introduction of the spike to the covariance, makes the problem much more complicated. This is made clear by Theorem \ref{thm:theorem3}, where in contrast to the isotropic feature case where initializing with some fixed correlation puts us immediately into the descent phase allowing for recovery with near linear sample complexity \cite{arous2021online}, with the introduction of the spike with any positive magnitude, a local minima appears around $m_1(X) = 0$ and hence with random initialization or initializing with some fixed correlation that is within the attractor region of the local optima, SGD tends to the local optima, in effect learning nothing and perhaps destroying the initial information.

We quickly point out that Theorem \ref{thm:theorem3} holds even with exponential data, stating that with the given stepsize, SGD does not recover the unknown parameter vector. We also note that the step-size specified in Theorem \ref{thm:theorem3} is in fact more general than in \ref{thm:theorem1} and \ref{thm:theorem2}. Hence, this is a reasonable range of stepsizes for which one would expect to solve the problem with sufficiently many samples.

\subsection{Meeting Assumptions}

We now take a moment to consider the assumptions in our theorems. Assumption \ref{assumption:Assumption B} requires the existence of some point $m^*$ such that within the rectangle defined by this point and the global optima $(m_1(X), m_2(X)) = (1,0)$, the population dynamics are well behaved, tending to the global optima at $(m_1(X), m_2(X)) = (1,0)$ in linear time. When it comes to the moment conditions on $f$ required to apply Theorems \ref{thm:theorem2} and \ref{thm:theorem3}, one can easily check (see Lemma~\ref{lemma:hermite} in the supplementary material) that the Hermite polynomials with degree $\geq 3$ satisfy them. %For a definition and nice introduction to the Hermite polynomials see \citet{patarroyo2019digression}.
% \begin{lemma}\label{lemma:hermite}
%     For all Hermite Polynomials with degree 3 or greater, $h_k(x), k \geq 3$: \[\mathbb{E}h_k''(g) = \mathbb{E}h_k'(g) = 0,\;\; \mathbb{E}\frac{\partial ^2}{\partial g^2}f(g)^2 > 0\] for $g \sim \mathcal{N}(0,1)$
% \end{lemma}
%This lemma provides a class of functions for which the assumptions of Theorems \ref{thm:theorem2} and \ref{thm:theorem3} apply. 
While this claim does not extend to all linear combinations of Hermite Polynomials, it can be extended to linear combinations of Hermite Polynomials of degrees 3 or greater, with the added constraint that any two coefficients in the Hermite expansion that are exactly 2 degrees apart, must have the same sign, i.e. $\mathbb{E}f(g)h_k(g)\mathbb{E}f(g)h_{k-2}(g) \geq 0$. This provides a class of functions which demonstrate our Theorems and thus the effects of pre-training.

\section{Transfer learning}

\label{finetune}

We also consider a related problem which we find lends itself better to the notion of transfer learning. We consider a related scenario under which we once again have labeled data $(y_i, a_i)_{i=1}^N$ according to a single layer network with known activation function $f$ and unknown parameter vector $v_0$. We assume that $f$ is differentiable almost everywhere with $f,f'$ of at most polynomial growth. However, we now consider the case of isotropic Gaussian features: $a_i \sim \mathcal{N}(0, I_d)$. Without the spike in the covariance, there is only one correlation variable of interest, namely $m_1(X)$ as previously defined. It can be shown that for $f$ differentiable almost everywhere and $f'$ of at most polynomial growth, the population loss $\Phi(X)$ can be expressed as $\phi(m_1(X)) \in C^1$, with $\phi'(x) < 0, \forall x \in (0,1)$ and further, the sample complexity for solving this problem with SGD is well understood \cite{arous2021online}.

To introduce the notion of transfer learning, we consider the scenario where we have access to some sequence of vectors $v^{(d)}$ with $ v^{(d)} \cdot v_0^{(d)}= \eta_{d}$. We may consider these correlated vectors $v^{(d)}$ to be a sequence of estimates of some vector correlated to $v_0$ which were obtained via SGD on some related task. For the sake of analysis we are not concerned with how these correlated vectors are obtained, only the benefit provided by having access to them for initializing SGD. We are again interested in the sample complexity as a function of dimension and how this complexity is affected by initializing with transfer learning in contrast to uniform random initializations.

Recall the information exponent from \cite{arous2021online}.

\begin{definition}\label{def:info}
    We say that a population loss $\phi$ has \textbf{information exponent} k if $\phi \in C^{k+1}([-1,1])$ and there exist $C, c > 0 $ such that:
$$\begin{cases}
\frac{d^\ell \phi}{dm^\ell}(0) = 0 & \text{for } 1 \leq \ell < k, \\
\frac{d^k \phi}{dm^k}(0) \leq -c < 0, \\
\left\| \frac{d^{k+1} \phi}{dm^{k+1}}(m) \right\|_\infty \leq C.
\end{cases}$$

\end{definition}

We now state a result concerning transfer learning with isotropic Gaussian features as described above. The details on sample complexity for this model are well understood from the work of \citet{arous2021online}, which allows us to easily analyze the effects of transfer learning.

\begin{theorem}\label{thm:TL}
     Let $k \geq 2$ be the information exponent of $\phi$. Let $v^{(d)} \cdot v_{0}^{(d)} = \eta_d = \theta(d^{-\zeta})$, with $\zeta \in [0,1/2)$. Then for spherical SGD with $N=\alpha  d$ steps with $\alpha \gg d^{2\zeta(k-2)} (\log d)^{2 \textbf{1}[\zeta > 0]}$ and $\alpha^{-1} \ll \delta \ll \alpha^{-1/2} $, $X_0 = v$, we have that: $m_1(X) \to 1$
    in probability as $d \to \infty$. Here $\textbf{1}[\zeta > 0]$ is the indicator function, taking value 0 if $\zeta = 0$ and 1 otherwise.

\end{theorem}

The proof of this theorem follows almost exactly from \cite{arous2021online}, noting that their arguments still hold under slightly different initializations. Contrasting this theorem with the results of \cite{arous2021online}, we notice that when $v^{(d)} \cdot v_{0}^{(d)} = \eta_d = O(d^{-\zeta})$ for some $\zeta \in (0, 1/2)$ and the information exponent is 3 or greater, we benefit from a polynomial sample reduction from $\alpha \gg d^{k-2}(\log d)^2$ to $\alpha \gg d^{2\zeta(k-2)} (\log d)^{2}$. Further in the case that $\zeta = 0$, i.e., we initialize with a fixed correlation independent of the dimension, we see that only $\alpha = \omega(1)$ is required, and hence we can recover in nearly linear sample complexity regardless of the information exponent. This offers a substantial polynomial reduction in the event that the information exponent is large. We do note however that even in the case the of the information exponent 2, we still benefit from a complexity reduction of a factor of $(\log d)^2$. In the case of information exponent 1, we do not benefit from transfer learning, however, in this case we already recover with nearly linear sample complexity, hence we have no need to perform transfer learning. We also note here that the reduction in sample complexity is the same as observed in Theorem 7 of \citet{mousavi2023gradient}.

\section{Examples}

In this section we show some examples of our theorems and provide simulations to empirically verify our claims in large finite dimensions.

\subsection{The Third Hermite Polynomial}

We will now show how one can apply our theorems from the past section. As noted in the past section, for all Hermite polynomials with degree 3 or greater, all assumptions are met for applying Theorems \ref{thm:theorem2} and \ref{thm:theorem1}, provided a large enough correlation between $v$ and $v_0$, ie for large enough $\eta_1$. Hence in order to apply our theorems to a specific problem set up with some Hermite polynomial in place of $f$ and some values of $\lambda, \eta_1$ we must simply verify whether or not $\eta_1$ is large enough. To do this, we must identify the region of Effective initializations for population flow and determine whether $(\eta_1, \eta_2)$ falls inside. 

Note that taking $f$ to be any polynomial function, we solve for $\Phi(X)$ explicitly, by expanding and computing the moments of Gaussian random variables. After computing the explicit loss, one can compute the spherical gradients with respect to $x_1$ and $x_2$ and analyze their signs in order to identify a value of $m^*$ to apply Assumption \ref{assumption:Assumption B}. Below we plot the phase diagram for population flow when $f(x) = x^3 - 3x$, the third Hermite polynomial. We identify a point $m^*$ to Apply assumption \ref{assumption:Assumption B}.

\subsection{Simulations}
\label{simulation_section}

\begin{figure*}[t!]
\centering
\begin{subfigure}[]

  \includegraphics[scale=0.4]{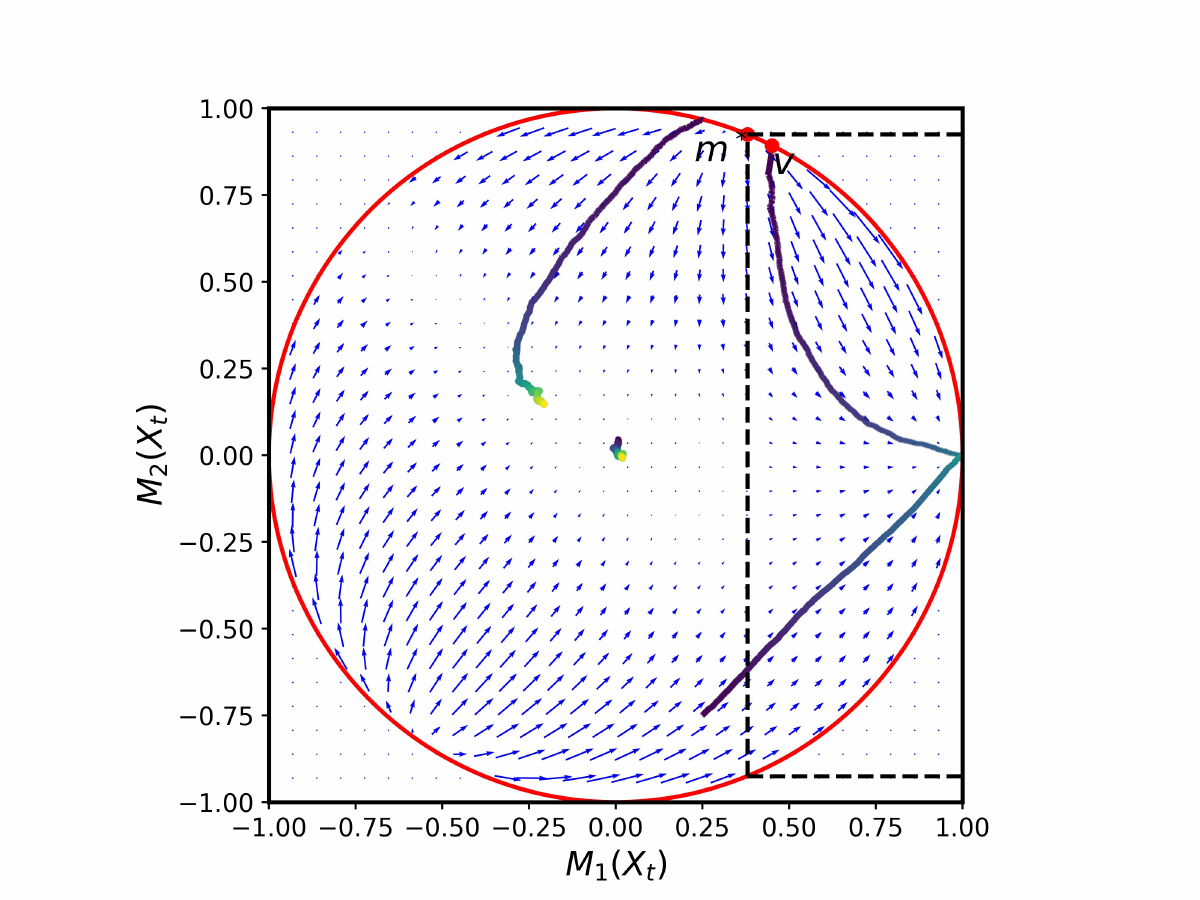} \label{fig:a}
\end{subfigure}
\begin{subfigure}[]

   \includegraphics[scale=0.4]{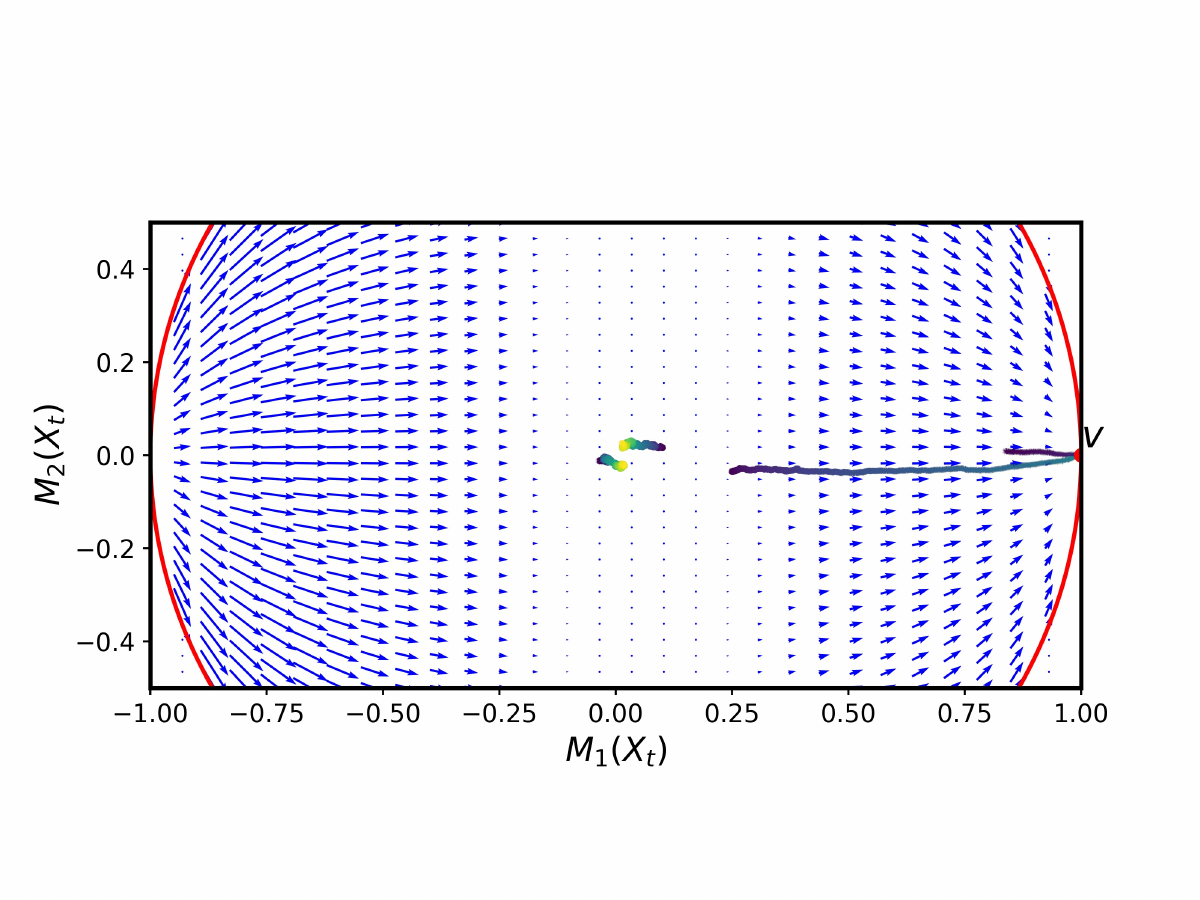} \label{fig:b}
\end{subfigure}  \\ 
\begin{subfigure}[]

     \includegraphics[scale=0.4]{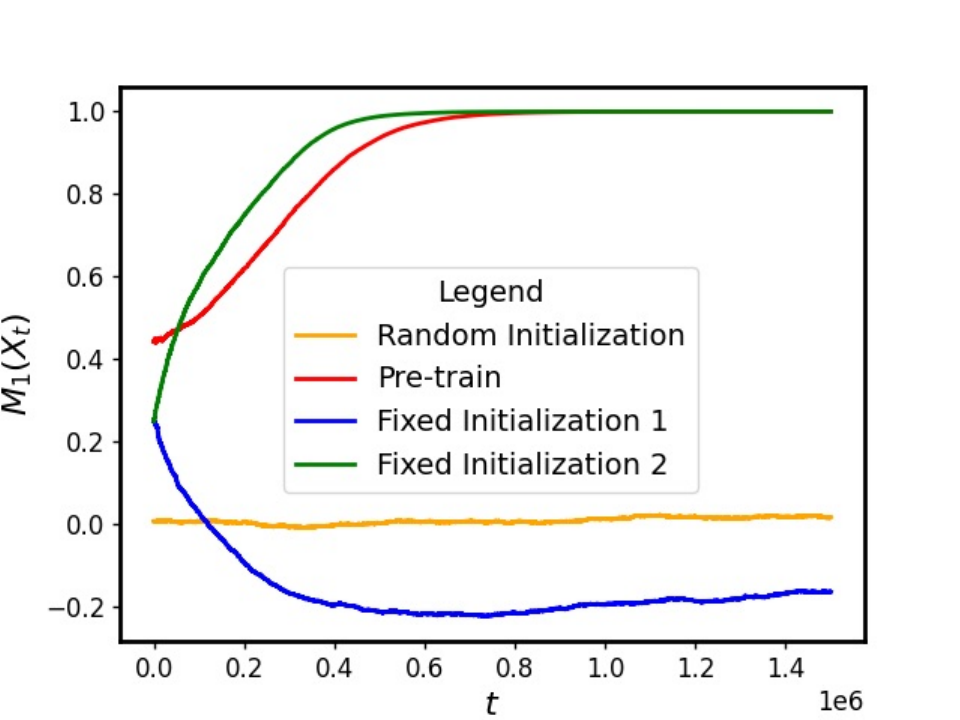} \label{fig:c}
\end{subfigure}
\begin{subfigure}[]

   \includegraphics[scale=0.4]{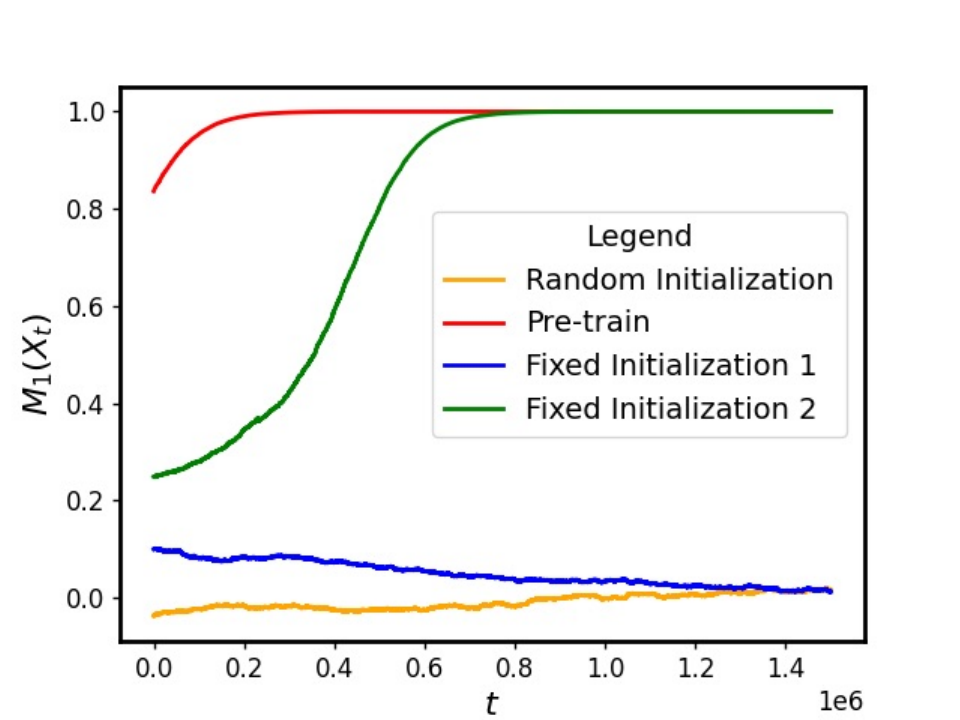} \label{fig:d}
\end{subfigure}
\caption{The figures on the left correspond to $\eta_1= 0.45, \lambda = 1$ and the right correspond $\eta_1 = 1, \lambda = 0.5$. Each figure displays 4 SGD runs. One random initialization, one initialization via pre-training with PCA and 2 fixed initializations. The left figures feature fixed initializations of $(m_1(X_0), m_2(X_0)) = (0.25, \sqrt{1-0.25^2})$ and $(m_1(X_0), m_2(X_0)) = (0.25, -0.75)$. We observe that pre-training results in finding the global optima at $(m_1(X_0), m_2(X_0)) = (1,0)$ and random initialization makes little to no progress. Fixing the value of $m_1(X_0)$, we also observe very different behaviors simply by varying the value of $m_2(X_0)$ highlighting the complexity that arises when introducing a second dimension to the population dynamics. In the case of $\eta_1=1$ the fixed initializations are $m_1(X_0) = 0.1$ and $m_1(X_0) = 0.25$. In addition to noticing the behavior of random initialization versus pre-training, we observe very different behavior for fixed initializations. For small enough $m_1(X_0)$ SGD tends towards the local optima at $m_1(X) = 0$ as suggested by Theorem \ref{thm:theorem3}. For larger enough $m_1(X_0)$, SGD tends towards the global optima.}
\label{fig:simulations}
\end{figure*}

We conduct a few simulations to empirically demonstrate our claims in finite dimensions. In the first simulation, we consider letting $f(x) = x^3 - 3x$ and setting $\lambda = 1, \eta_1 = 0.45$. We then conduct SGD from both random initializations and from estimates of $v$ obtained via PCA. We use dimension $d = 1000$ and let SGD run for $\frac{3}{2}d^2 = 1,500,000$ steps of size $\frac{1}{10d^2} = (10,000,000)^{-1}$. We select the parameters such that we would expect to be able to recover the true parameter vector from a random initialization had we been in the case $\lambda = 0$. We determine this scaling based on the results of \citet{arous2021online} and some experimenting. See Figure \ref{fig:simulations}.

We also perform simulations under the setting $\lambda = 0.5$ and $\eta_1 = 1$ as in Theorem \ref{thm:theorem3}. We then perform SGD with the dimension, step size and number as steps as given above, only we consider initializing uniform randomly, conditional on fixing the correlation $m_1(X_0) = 0.1$. See Figure \ref{fig:simulations}.
%%See Appendix \ref{appendix_simulations}.

%\begin{figure}[ht]
%\begin{center}
%\centerline{\includegraphics[width=\columnwidth]{phase_diagram.pdf}}
%\caption{A phase diagram for population flow when $f(x) = x^3 - 3x$ (the third Hermite polynomial), $\lambda = 1, \eta_1=0.45$. We plot the point $m^*$ for which Assumption \ref{assumption:Assumption B} applies, and it's corresponding rectangle. We see that when initialized near $v = (\eta_1, \eta_2)$ as is done when we initialize with PCA, the initial value for population dynamics falls inside the effective region as seen by the $m^*$-rectangle.}
%\label{phase_diagram}
%\end{center}
%\end{figure}

\section{Proof Ideas}

We make use of the `bounding flows' approach from \citet{ben2020bounding}, \citet{arous2020algorithmic}, and \citet{arous2021online}. This approach was applied to Single-Index Models in \cite{arous2021online}. The key difference here is that  the populations dynamics here cannot be reduced to a 1-dimensional correlation variable but a 2-dimensional vector of correlations. As such more delicate analysis of the phase portrait and the martingale fluctuations are involved.  %These techniques involve using discrete analogues of differential inequalities from Gronwall \cite{gross1967gronwall} and Bihari-LaSalle (see Appendix~\ref{bihari_appendix}) to bound the distance between SGD and population flow. Additionally, making use of Doob's Inequality \cite{williams1991probability} to control the martingale term which is seen as the accumulated sample-wise errors, where sample-wise errors are the difference in the gradient of the (random) loss and it's expectation. 

\subsection{Theorem \ref{thm:theorem1}}

Our first main theorem, provides a sufficient condition to check when pre-training for initializing SGD can recover the unknown parameter vector in almost linear time. The proof of  has two main components. First note that under Assumption \ref{assumption:Assumption B}, there exists a rectangle that, when initialized within, the population dynamics will find the global optima. Next Lemma \ref{lemma:2} shows that when initializing with some fixed initial correlations, SGD behaves like the population dynamics, which under Assumption \ref{assumption:Assumption B} converge to the global optima when initialized in the rectangle. The second part of the proof of Theorem \ref{thm:theorem1} is simply applying a few well-known facts regarding PCA in high dimensions (see Appendix \ref{appendix:b}). These facts along with the assumption on the strength of the correlation of the spike to the unknown parameter vector, tell us that when initializing with pre-training, we find ourselves in the region of effective initializations as given by Assumption \ref{assumption:Assumption B} and hence we can recover the unknown parameter vector in approximately linear time.

\subsection{Theorem \ref{thm:theorem2}}

 We consider the Taylor expansion of the population loss in the two correlation variables of interest around 0. Recall that random uniform initializations yield initial correlations on the order of $m_1(X_0), m_2(X_0) = O(\frac{1}{\sqrt{d}})$ \cite{Vershynin_2018}. Under the Assumptions on $f$ of Theorems \ref{thm:theorem2} and \ref{thm:theorem3} we have the following system for the population dynamics:

\begin{align*}
    &\frac{\partial \phi(x_{1}, x_{2})}{\partial x_{1}} =  2\lambda \eta_{1}^2cx_1 + 2\lambda \eta_{1} \eta_{2}cx_{2} + O(\|(x_1, x_2)\|_2^2)\\& \frac{\partial \phi(x_{1}, x_{2})}{\partial x_{2}} =  2\lambda \eta_{2}^2 cx_{2} + 2\lambda \eta_{1} \eta_{2}cx_{1} + O(\|(x_1, x_2)\|_2^2) 
\end{align*}

for some positive constant c. We remind the reader here of the definition $(x_1, x_2) = (m_1(X), m_2(X))$. Analyzing the linearized system, we see that the first-order terms are orthogonal to (and point towards) the line $L = \{ (x_1, x_2) : x_1 = -\frac{\eta_2}{\eta_1}x_2\}$, and are equal to 0 on $L$. Letting $T_L^\perp(x_1, x_2)$ measure the distance of $(x_1, x_2)$ to $L$, we have that the first terms exceed higher order terms when outside the set $\mathcal{C} = \{ (x_1, x_2): T_L^\perp(x_1, x_2) > c_1\|(x_1, x_2)\|_2^2\}$ for some constant $c_1$. This set provides a cusp surrounding the line $L$. To prove our result, we carefully construct stopping times in order to observe the process over specific regions of the space, such as the Cartesian quadrants and positioning relative to the set $\mathcal{C}$.
We show that in quadrants 1 and 3, the process tends to quadrant 2 or quadrant 4. Once in these quadrants we bound the distance of the process to the line $L$, ultimately ensuring the process gets close to $L$. Once this happens we show that the given sample complexity is not sufficient to leave some fixed ball around the origin. The proof of this theorem is the most involved. While this idea can be understood at the population level, the extension to SGD requires controlling the variance of the martingale along various important directions with martingale inequalities.

\subsection{Theorem \ref{thm:theorem3}}

When $\eta_1 = 1$ (and hence $\eta_2 = 0$) as in Theorem \ref{thm:theorem3}, the system no longer depends on $x_2$ (Recall the definition of $\phi$ in the start of section \ref{main}). We then have the following 1-d system:
\[
\frac{\partial \phi(x_{1}, x_{2})}{\partial x_{1}} =  2\lambda \eta_{1}^2cx_1 + O(\|(x_1, x_2)\|_2^2)
\]
From this system we see there is a local optima at $m_1(X) = 0$, whose attractor region is fixed and not dependent on the dimension. Hence the remainder of the proof of Theorem \ref{thm:theorem3} is showing that in the high dimensional limit, the randomness of SGD is insufficient to escape the attractor region in the given scaling regime. 

\section{Conclusion}

In this paper we consider natural statistical models for which one can analyze the effects of pre-training and transfer learning. Namely single-index models with Gaussian features with spiked covariance and isotropic covariance. We analyze the ability to recover the unknown parameter vector in these models using stochastic gradient descent and the required sample complexity in the high dimensional regime, from both random initializations and with pre-training / transfer learning. In both scenarios we prove polynomial separation in the sample complexity as a function of the dimension required to solve, for a class of functions which contains the Hermite polynomials of degree 3 or greater. We also highlight the complexity of analyzing recovery under a single-index model with Gaussian features and spiked covariance, by highlighting a simple case (the case where the spike vector is equal to the parameter vector) in which a local optima arises and traps random initializations. This paper contributes to the growing body of work attempting to add theoretical justification for common practices of pre-training and transfer learning.

%\newpage

\section*{Acknowledgments}

A.J.\ acknowledges the support of the Natural Sciences and Engineering Research Council of Canada (NSERC), the Canada Research Chairs programme, and the Ontario Research Fund. Cette recherche a \'et\'e enterprise gr\^ace, en partie, au 
soutien financier du Conseil de Recherches en Sciences Naturelles et en G\'enie du Canada (CRSNG),  [RGPIN-2020-04597, DGECR-2020-00199], et du Programme des chaires de recherche du Canada.  S.S. gratefully acknowledges support from NSF (DMS CAREER 2239234), ONR (N00014-23-1-2489) and AFOSR (FA9950-23-1-0429). T.J acknowledges the support of the Natural Sciences and Engineering Research Council of Canada (NSERC). 

\section*{Impact Statement}

This paper focuses on advancing the theoretical foundations of Machine Learning, and we do not anticipate significant societal consequences from our work.

\bibliography{example_paper}

\begin{thebibliography}{67}
\providecommand{\natexlab}[1]{#1}
\providecommand{\url}[1]{\texttt{#1}}
\expandafter\ifx\csname urlstyle\endcsname\relax
  \providecommand{\doi}[1]{doi: #1}\else
  \providecommand{\doi}{doi: \begingroup \urlstyle{rm}\Url}\fi

\bibitem[Abbe et~al.(2023)Abbe, Adsera, and Misiakiewicz]{abbe2023sgd}
Abbe, E., Adsera, E.~B., and Misiakiewicz, T.
\newblock Sgd learning on neural networks: leap complexity and saddle-to-saddle dynamics.
\newblock In \emph{The Thirty Sixth Annual Conference on Learning Theory}, pp.\  2552--2623. PMLR, 2023.

\bibitem[Achiam et~al.(2023)Achiam, Adler, Agarwal, Ahmad, Akkaya, Aleman, Almeida, Altenschmidt, Altman, Anadkat, et~al.]{achiam2023gpt}
Achiam, J., Adler, S., Agarwal, S., Ahmad, L., Akkaya, I., Aleman, F.~L., Almeida, D., Altenschmidt, J., Altman, S., Anadkat, S., et~al.
\newblock Gpt-4 technical report.
\newblock \emph{arXiv preprint arXiv:2303.08774}, 2023.

\bibitem[Albuquerque et~al.(2019)Albuquerque, Monteiro, Darvishi, Falk, and Mitliagkas]{albuquerque2019generalizing}
Albuquerque, I., Monteiro, J., Darvishi, M., Falk, T.~H., and Mitliagkas, I.
\newblock Generalizing to unseen domains via distribution matching.
\newblock \emph{arXiv preprint arXiv:1911.00804}, 2019.

\bibitem[Arnaboldi et~al.(2024)Arnaboldi, Dandi, Krzakala, Pesce, and Stephan]{arnaboldi2024repetita}
Arnaboldi, L., Dandi, Y., Krzakala, F., Pesce, L., and Stephan, L.
\newblock Repetita iuvant: Data repetition allows sgd to learn high-dimensional multi-index functions.
\newblock \emph{arXiv preprint arXiv:2405.15459}, 2024.

\bibitem[Arora et~al.(2019)Arora, Khandeparkar, Khodak, Plevrakis, and Saunshi]{arora2019theoretical}
Arora, S., Khandeparkar, H., Khodak, M., Plevrakis, O., and Saunshi, N.
\newblock A theoretical analysis of contrastive unsupervised representation learning.
\newblock \emph{arXiv preprint arXiv:1902.09229}, 2019.

\bibitem[Azar \& Nadler(2025)Azar and Nadler]{azar2024semi}
Azar, E. and Nadler, B.
\newblock Semi-supervised sparse gaussian classification: Provable benefits of unlabeled data.
\newblock \emph{Advances in Neural Information Processing Systems}, 2025.

\bibitem[Ba et~al.(2024)Ba, Erdogdu, Suzuki, Wang, and Wu]{ba2024learning}
Ba, J., Erdogdu, M.~A., Suzuki, T., Wang, Z., and Wu, D.
\newblock Learning in the presence of low-dimensional structure: a spiked random matrix perspective.
\newblock \emph{Advances in Neural Information Processing Systems}, 36, 2024.

\bibitem[Bai \& Silverstein(2010)Bai and Silverstein]{bai2010spectral}
Bai, Z. and Silverstein, J.~W.
\newblock \emph{Spectral analysis of large dimensional random matrices}, volume~20.
\newblock Springer, 2010.

\bibitem[Baik et~al.(2005)Baik, Ben~Arous, and P{\'e}ch{\'e}]{baik2005phase}
Baik, J., Ben~Arous, G., and P{\'e}ch{\'e}, S.
\newblock Phase transition of the largest eigenvalue for nonnull complex sample covariance matrices.
\newblock 2005.

\bibitem[Baldi \& Hornik(1989)Baldi and Hornik]{baldi1989neural}
Baldi, P. and Hornik, K.
\newblock Neural networks and principal component analysis: Learning from examples without local minima.
\newblock \emph{Neural networks}, 2\penalty0 (1):\penalty0 53--58, 1989.

\bibitem[Bandeira et~al.(2020)Bandeira, Singer, and Strohmer]{MDS}
Bandeira, A.~S., Singer, A., and Strohmer, T.
\newblock \emph{Mathematics of Data Science}.
\newblock 2020.
\newblock URL \url{https://people.math.ethz.ch/~abandeira/BandeiraSingerStrohmer-MDS-draft.pdf}.

\bibitem[Barbier et~al.(2019)Barbier, Krzakala, Macris, Miolane, and Zdeborov{\'a}]{barbier2019optimal}
Barbier, J., Krzakala, F., Macris, N., Miolane, L., and Zdeborov{\'a}, L.
\newblock Optimal errors and phase transitions in high-dimensional generalized linear models.
\newblock \emph{Proceedings of the National Academy of Sciences}, 116\penalty0 (12):\penalty0 5451--5460, 2019.

\bibitem[Ben~Arous et~al.(2020{\natexlab{a}})Ben~Arous, Gheissari, and Jagannath]{arous2020algorithmic}
Ben~Arous, G., Gheissari, R., and Jagannath, A.
\newblock Algorithmic thresholds for tensor pca.
\newblock \emph{The Annals of Probability}, 48\penalty0 (4):\penalty0 2052--2087, 2020{\natexlab{a}}.

\bibitem[Ben~Arous et~al.(2020{\natexlab{b}})Ben~Arous, Gheissari, and Jagannath]{ben2020bounding}
Ben~Arous, G., Gheissari, R., and Jagannath, A.
\newblock Bounding flows for spherical spin glass dynamics.
\newblock \emph{Communications in Mathematical Physics}, 373:\penalty0 1011--1048, 2020{\natexlab{b}}.

\bibitem[Ben~Arous et~al.(2021)Ben~Arous, Gheissari, and Jagannath]{arous2021online}
Ben~Arous, G., Gheissari, R., and Jagannath, A.
\newblock Online stochastic gradient descent on non-convex losses from high-dimensional inference.
\newblock \emph{Journal of Machine Learning Research}, 22\penalty0 (106):\penalty0 1--51, 2021.

\bibitem[Ben-David et~al.(2010)Ben-David, Blitzer, Crammer, Kulesza, Pereira, and Vaughan]{ben2010theory}
Ben-David, S., Blitzer, J., Crammer, K., Kulesza, A., Pereira, F., and Vaughan, J.~W.
\newblock A theory of learning from different domains.
\newblock \emph{Machine learning}, 79:\penalty0 151--175, 2010.

\bibitem[Bietti et~al.(2023)Bietti, Bruna, and Pillaud-Vivien]{bietti2023learning}
Bietti, A., Bruna, J., and Pillaud-Vivien, L.
\newblock On learning gaussian multi-index models with gradient flow.
\newblock \emph{arXiv preprint arXiv:2310.19793}, 2023.

\bibitem[Bourlard \& Kamp(1988)Bourlard and Kamp]{bourlard1988auto}
Bourlard, H. and Kamp, Y.
\newblock Auto-association by multilayer perceptrons and singular value decomposition.
\newblock \emph{Biological cybernetics}, 59\penalty0 (4):\penalty0 291--294, 1988.

\bibitem[Brown(2020)]{brown2020language}
Brown, T.~B.
\newblock Language models are few-shot learners.
\newblock \emph{arXiv preprint arXiv:2005.14165}, 2020.

\bibitem[Candes et~al.(2015)Candes, Li, and Soltanolkotabi]{candes2015phase}
Candes, E.~J., Li, X., and Soltanolkotabi, M.
\newblock Phase retrieval via wirtinger flow: Theory and algorithms.
\newblock \emph{IEEE Transactions on Information Theory}, 61\penalty0 (4):\penalty0 1985--2007, 2015.

\bibitem[Damian et~al.(2022)Damian, Lee, and Soltanolkotabi]{damian2022neural}
Damian, A., Lee, J., and Soltanolkotabi, M.
\newblock Neural networks can learn representations with gradient descent.
\newblock In \emph{Conference on Learning Theory}, pp.\  5413--5452. PMLR, 2022.

\bibitem[Damian et~al.(2024{\natexlab{a}})Damian, Nichani, Ge, and Lee]{damian2024smoothing}
Damian, A., Nichani, E., Ge, R., and Lee, J.~D.
\newblock Smoothing the landscape boosts the signal for sgd: Optimal sample complexity for learning single index models.
\newblock \emph{Advances in Neural Information Processing Systems}, 36, 2024{\natexlab{a}}.

\bibitem[Damian et~al.(2024{\natexlab{b}})Damian, Pillaud-Vivien, Lee, and Bruna]{damian2024computational}
Damian, A., Pillaud-Vivien, L., Lee, J., and Bruna, J.
\newblock Computational-statistical gaps in gaussian single-index models.
\newblock In \emph{The Thirty Seventh Annual Conference on Learning Theory}, pp.\  1262--1262. PMLR, 2024{\natexlab{b}}.

\bibitem[Dandi et~al.(2024)Dandi, Troiani, Arnaboldi, Pesce, Zdeborov{\'a}, and Krzakala]{dandi2024benefits}
Dandi, Y., Troiani, E., Arnaboldi, L., Pesce, L., Zdeborov{\'a}, L., and Krzakala, F.
\newblock The benefits of reusing batches for gradient descent in two-layer networks: Breaking the curse of information and leap exponents.
\newblock \emph{International Conference on Machine Learning}, 2024.

\bibitem[David et~al.(2010)David, Lu, Luu, and P{\'a}l]{david2010impossibility}
David, S.~B., Lu, T., Luu, T., and P{\'a}l, D.
\newblock Impossibility theorems for domain adaptation.
\newblock In \emph{Proceedings of the Thirteenth International Conference on Artificial Intelligence and Statistics}, pp.\  129--136. JMLR Workshop and Conference Proceedings, 2010.

\bibitem[Devlin et~al.(2019)Devlin, Chang, Lee, and Toutanova]{devlin2019bert}
Devlin, J., Chang, M.-W., Lee, K., and Toutanova, K.
\newblock Bert: Pre-training of deep bidirectional transformers for language understanding.
\newblock In \emph{Proceedings of the 2019 Conference of the North American Chapter of the Association for Computational Linguistics: Human Language Technologies}, volume~1, pp.\  4171--4186, Minneapolis, Minnesota, 2019. Association for Computational Linguistics.

\bibitem[Dudeja \& Hsu(2018)Dudeja and Hsu]{dudeja2018learning}
Dudeja, R. and Hsu, D.
\newblock Learning single-index models in gaussian space.
\newblock In \emph{Conference On Learning Theory}, pp.\  1887--1930. PMLR, 2018.

\bibitem[Ganin et~al.(2016)Ganin, Ustinova, Ajakan, Germain, Larochelle, Laviolette, March, and Lempitsky]{ganin2016domain}
Ganin, Y., Ustinova, E., Ajakan, H., Germain, P., Larochelle, H., Laviolette, F., March, M., and Lempitsky, V.
\newblock Domain-adversarial training of neural networks.
\newblock \emph{Journal of machine learning research}, 17\penalty0 (59):\penalty0 1--35, 2016.

\bibitem[Gerace et~al.(2022)Gerace, Saglietti, Mannelli, Saxe, and Zdeborov{\'a}]{gerace2022probing}
Gerace, F., Saglietti, L., Mannelli, S.~S., Saxe, A., and Zdeborov{\'a}, L.
\newblock Probing transfer learning with a model of synthetic correlated datasets.
\newblock \emph{Machine Learning: Science and Technology}, 3\penalty0 (1):\penalty0 015030, 2022.

\bibitem[Gross(1967)]{gross1967gronwall}
Gross, L.
\newblock Gronwall's inequality and its applications.
\newblock \emph{The Journal of Mathematical Analysis and Applications}, 20\penalty0 (3):\penalty0 359--370, 1967.

\bibitem[Hanneke \& Kpotufe(2019)Hanneke and Kpotufe]{hanneke2019value}
Hanneke, S. and Kpotufe, S.
\newblock On the value of target data in transfer learning.
\newblock \emph{Advances in Neural Information Processing Systems}, 32, 2019.

\bibitem[He et~al.(2017)He, Gkioxari, Doll{\'a}r, and Girshick]{he2017mask}
He, K., Gkioxari, G., Doll{\'a}r, P., and Girshick, R.
\newblock Mask r-cnn.
\newblock In \emph{Proceedings of the IEEE international conference on computer vision}, pp.\  2961--2969, 2017.

\bibitem[Heckman(1979)]{heckman1979sample}
Heckman, J.~J.
\newblock Sample selection bias as a specification error.
\newblock \emph{Econometrica: Journal of the econometric society}, pp.\  153--161, 1979.

\bibitem[Huang et~al.(2006)Huang, Gretton, Borgwardt, Sch{\"o}lkopf, and Smola]{huang2006correcting}
Huang, J., Gretton, A., Borgwardt, K., Sch{\"o}lkopf, B., and Smola, A.
\newblock Correcting sample selection bias by unlabeled data.
\newblock \emph{Advances in neural information processing systems}, 19, 2006.

\bibitem[Lee et~al.(2021)Lee, Lei, Saunshi, and Zhuo]{lee2021predicting}
Lee, J.~D., Lei, Q., Saunshi, N., and Zhuo, J.
\newblock Predicting what you already know helps: Provable self-supervised learning.
\newblock \emph{Advances in Neural Information Processing Systems}, 34:\penalty0 309--323, 2021.

\bibitem[Lee et~al.(2024)Lee, Oko, Suzuki, and Wu]{lee2024neural}
Lee, J.~D., Oko, K., Suzuki, T., and Wu, D.
\newblock Neural network learns low-dimensional polynomials with sgd near the information-theoretic limit.
\newblock \emph{arXiv preprint arXiv:2406.01581}, 2024.

\bibitem[Long et~al.(2017)Long, Zhu, Wang, and Jordan]{long2017deep}
Long, M., Zhu, H., Wang, J., and Jordan, M.~I.
\newblock Deep transfer learning with joint adaptation networks.
\newblock In \emph{International conference on machine learning}, pp.\  2208--2217. PMLR, 2017.

\bibitem[Lu \& Li(2020)Lu and Li]{lu2020phase}
Lu, Y.~M. and Li, G.
\newblock Phase transitions of spectral initialization for high-dimensional non-convex estimation.
\newblock \emph{Information and Inference: A Journal of the IMA}, 9\penalty0 (3):\penalty0 507--541, 2020.

\bibitem[Maillard et~al.(2020{\natexlab{a}})Maillard, Ben~Arous, and Biroli]{maillard2019landscape}
Maillard, A., Ben~Arous, G., and Biroli, G.
\newblock Landscape complexity for the empirical risk of generalized linear models.
\newblock In Lu, J. and Ward, R. (eds.), \emph{Proceedings of The First Mathematical and Scientific Machine Learning Conference}, volume 107 of \emph{Proceedings of Machine Learning Research}, pp.\  287--327. PMLR, 20--24 Jul 2020{\natexlab{a}}.
\newblock URL \url{https://proceedings.mlr.press/v107/maillard20a.html}.

\bibitem[Maillard et~al.(2020{\natexlab{b}})Maillard, Loureiro, Krzakala, and Zdeborov{\'a}]{maillard2020phase}
Maillard, A., Loureiro, B., Krzakala, F., and Zdeborov{\'a}, L.
\newblock Phase retrieval in high dimensions: Statistical and computational phase transitions.
\newblock \emph{Advances in Neural Information Processing Systems}, 33:\penalty0 11071--11082, 2020{\natexlab{b}}.

\bibitem[Maity et~al.(2022)Maity, Sun, and Banerjee]{maity2022minimax}
Maity, S., Sun, Y., and Banerjee, M.
\newblock Minimax optimal approaches to the label shift problem in non-parametric settings.
\newblock \emph{Journal of Machine Learning Research}, 23\penalty0 (346):\penalty0 1--45, 2022.

\bibitem[Mousavi-Hosseini et~al.(2023)Mousavi-Hosseini, Wu, Suzuki, and Erdogdu]{mousavi2023gradient}
Mousavi-Hosseini, A., Wu, D., Suzuki, T., and Erdogdu, M.~A.
\newblock Gradient-based feature learning under structured data.
\newblock \emph{Advances in Neural Information Processing Systems}, 36:\penalty0 71449--71485, 2023.

\bibitem[Nguyen(2021)]{nguyen2021analysis}
Nguyen, P.-M.
\newblock Analysis of feature learning in weight-tied autoencoders via the mean field lens.
\newblock \emph{arXiv preprint arXiv:2102.08373}, 2021.

\bibitem[Pan \& Yang(2009)Pan and Yang]{pan2009survey}
Pan, S.~J. and Yang, Q.
\newblock A survey on transfer learning.
\newblock \emph{IEEE Transactions on knowledge and data engineering}, 22\penalty0 (10):\penalty0 1345--1359, 2009.

\bibitem[Pesce et~al.(2023)Pesce, Krzakala, Loureiro, and Stephan]{pesce2023gaussian}
Pesce, L., Krzakala, F., Loureiro, B., and Stephan, L.
\newblock Are gaussian data all you need? the extents and limits of universality in high-dimensional generalized linear estimation.
\newblock In \emph{International Conference on Machine Learning}, pp.\  27680--27708. PMLR, 2023.

\bibitem[Qui{\~n}onero-Candela et~al.(2022)Qui{\~n}onero-Candela, Sugiyama, Schwaighofer, and Lawrence]{quinonero2022dataset}
Qui{\~n}onero-Candela, J., Sugiyama, M., Schwaighofer, A., and Lawrence, N.~D.
\newblock \emph{Dataset shift in machine learning}.
\newblock Mit Press, 2022.

\bibitem[Radford et~al.(2019)Radford, Wu, Child, Luan, Amodei, Sutskever, et~al.]{radford2019language}
Radford, A., Wu, J., Child, R., Luan, D., Amodei, D., Sutskever, I., et~al.
\newblock Language models are unsupervised multitask learners.
\newblock \emph{OpenAI blog}, 1\penalty0 (8):\penalty0 9, 2019.

\bibitem[Ren \& Lee(2024)Ren and Lee]{ren2024learning}
Ren, Y. and Lee, J.~D.
\newblock Learning orthogonal multi-index models: A fine-grained information exponent analysis.
\newblock \emph{arXiv preprint arXiv:2410.09678}, 2024.

\bibitem[Sagawa et~al.(2019)Sagawa, Koh, Hashimoto, and Liang]{sagawa2019distributionally}
Sagawa, S., Koh, P.~W., Hashimoto, T.~B., and Liang, P.
\newblock Distributionally robust neural networks for group shifts: On the importance of regularization for worst-case generalization.
\newblock \emph{In International Conference on Learning Representations (ICLR)}, 2019.

\bibitem[Schneider et~al.(2019)Schneider, Baevski, Collobert, and Auli]{schneider2019wav2vec}
Schneider, S., Baevski, A., Collobert, R., and Auli, M.
\newblock wav2vec: Unsupervised pre-training for speech recognition.
\newblock \emph{arXiv preprint arXiv:1904.05862}, 2019.

\bibitem[Shimodaira(2000)]{shimodaira2000improving}
Shimodaira, H.
\newblock Improving predictive inference under covariate shift by weighting the log-likelihood function.
\newblock \emph{Journal of statistical planning and inference}, 90\penalty0 (2):\penalty0 227--244, 2000.

\bibitem[Storkey(2008)]{storkey2008training}
Storkey, A.
\newblock When training and test sets are different: characterizing learning transfer.
\newblock 2008.

\bibitem[Sun et~al.(2018)Sun, Qu, and Wright]{sun2018geometric}
Sun, J., Qu, Q., and Wright, J.
\newblock A geometric analysis of phase retrieval.
\newblock \emph{Foundations of Computational Mathematics}, 18:\penalty0 1131--1198, 2018.

\bibitem[Tachet~des Combes et~al.(2020)Tachet~des Combes, Zhao, Wang, and Gordon]{tachet2020domain}
Tachet~des Combes, R., Zhao, H., Wang, Y.-X., and Gordon, G.~J.
\newblock Domain adaptation with conditional distribution matching and generalized label shift.
\newblock \emph{Advances in Neural Information Processing Systems}, 33:\penalty0 19276--19289, 2020.

\bibitem[Tosh et~al.(2021{\natexlab{a}})Tosh, Krishnamurthy, and Hsu]{tosh2021contrastive}
Tosh, C., Krishnamurthy, A., and Hsu, D.
\newblock Contrastive estimation reveals topic posterior information to linear models.
\newblock \emph{Journal of Machine Learning Research}, 22\penalty0 (281):\penalty0 1--31, 2021{\natexlab{a}}.

\bibitem[Tosh et~al.(2021{\natexlab{b}})Tosh, Krishnamurthy, and Hsu]{tosh2021contrastive2}
Tosh, C., Krishnamurthy, A., and Hsu, D.
\newblock Contrastive learning, multi-view redundancy, and linear models.
\newblock In \emph{Algorithmic Learning Theory}, pp.\  1179--1206. PMLR, 2021{\natexlab{b}}.

\bibitem[Vershynin(2018)]{Vershynin_2018}
Vershynin, R.
\newblock \emph{High-Dimensional Probability: An Introduction with Applications in Data Science}.
\newblock Cambridge Series in Statistical and Probabilistic Mathematics. Cambridge University Press, 2018.

\bibitem[Wang et~al.(2016)Wang, Kurth-Nelson, Tirumala, Soyer, Leibo, Munos, Blundell, Kumaran, and Botvinick]{wang2016learning}
Wang, J.~X., Kurth-Nelson, Z., Tirumala, D., Soyer, H., Leibo, J.~Z., Munos, R., Blundell, C., Kumaran, D., and Botvinick, M.
\newblock Learning to reinforcement learn.
\newblock \emph{arXiv preprint arXiv:1611.05763}, 2016.

\bibitem[Wang \& Schneider(2015)Wang and Schneider]{wang2015generalization}
Wang, X. and Schneider, J.~G.
\newblock Generalization bounds for transfer learning under model shift.
\newblock In \emph{UAI}, pp.\  922--931, 2015.

\bibitem[Wang et~al.(2014)Wang, Huang, and Schneider]{wang2014active}
Wang, X., Huang, T.-K., and Schneider, J.
\newblock Active transfer learning under model shift.
\newblock In \emph{International Conference on Machine Learning}, pp.\  1305--1313. PMLR, 2014.

\bibitem[Wei et~al.(2021)Wei, Xie, and Ma]{wei2021pretrained}
Wei, C., Xie, S.~M., and Ma, T.
\newblock Why do pretrained language models help in downstream tasks? an analysis of head and prompt tuning.
\newblock \emph{Advances in Neural Information Processing Systems}, 34:\penalty0 16158--16170, 2021.

\bibitem[Williams(1991)]{williams1991probability}
Williams, D.
\newblock \emph{Probability with Martingales}.
\newblock Cambridge University Press, 1991.

\bibitem[Wu et~al.(2019)Wu, Winston, Kaushik, and Lipton]{wu2019domain}
Wu, Y., Winston, E., Kaushik, D., and Lipton, Z.
\newblock Domain adaptation with asymmetrically-relaxed distribution alignment.
\newblock In \emph{International conference on machine learning}, pp.\  6872--6881. PMLR, 2019.

\bibitem[Zhai et~al.(2023)Zhai, Liu, Risteski, Kolter, and Ravikumar]{zhai2023understanding}
Zhai, R., Liu, B., Risteski, A., Kolter, Z., and Ravikumar, P.
\newblock Understanding augmentation-based self-supervised representation learning via rkhs approximation and regression.
\newblock \emph{International Conference on Machine Learning}, 2023.

\bibitem[Zhang \& Hashimoto(2021)Zhang and Hashimoto]{zhang2021inductive}
Zhang, T. and Hashimoto, T.
\newblock On the inductive bias of masked language modeling: From statistical to syntactic dependencies.
\newblock \emph{arXiv preprint arXiv:2104.05694}, 2021.

\bibitem[Zhao et~al.(2019)Zhao, Des~Combes, Zhang, and Gordon]{zhao2019learning}
Zhao, H., Des~Combes, R.~T., Zhang, K., and Gordon, G.
\newblock On learning invariant representations for domain adaptation.
\newblock In \emph{International conference on machine learning}, pp.\  7523--7532. PMLR, 2019.

\bibitem[Zweig et~al.(2024)Zweig, Pillaud-Vivien, and Bruna]{zweig2024single}
Zweig, A., Pillaud-Vivien, L., and Bruna, J.
\newblock On single-index models beyond gaussian data.
\newblock \emph{Advances in Neural Information Processing Systems}, 36, 2024.

\end{thebibliography}
\bibliographystyle{icml2025}

% APPENDIX
%%%%%%%%%%%%%%%%%%%%%%%%%%%%%%%%%%%%%%%%%%%%%%%%%%%%%%%%%%%%%%%%%%%%%%%%%%%%%%%
%%%%%%%%%%%%%%%%%%%%%%%%%%%%%%%%%%%%%%%%%%%%%%%%%%%%%%%%%%%%%%%%%%%%%%%%%%%%%%%
%\newpage
\appendix
\onecolumn

\section{Proofs of Main Results}
\label{appendix}

For convenience we define the sample wise error as $H(X,y) = \mathcal{L}(X,y) -\Phi(X)$. We first prove a lemma which is used throughout our proofs.

\begin{lemma}\label{lemma:assumption_B}
    There exist constants $C_{1}, C_{2} > 0$ such that the following moment bounds hold uniformly in $d$:
\begin{align}
    &\sup_{x,\theta_{d} \in \mathbb{S}^{d-1}} \mathbb{E}[(\nabla H_{d}(x, y)\cdot \theta_{d})^2] \leq C_{1} \nonumber\\
&\sup_{x \in \mathbb{S}^{d-1}} \mathbb{E}[\| \nabla H_{d}(x, y)\|_{2}^{4 + C_{2}}] \leq C_{1} d^{(4 + C_{2})/2}
\nonumber\end{align}
\end{lemma}

\begin{proof}
    See the proof of proposition B.1 in \cite{arous2021online} and not that similar arguments apply when the features have spiked covariance.
\end{proof}

\subsection{Proof of Theorem \ref{thm:theorem1}. }

We first prove Theorem \ref{thm:theorem1}.

\begingroup
\renewcommand{\thetheorem}{\ref{thm:theorem1}} % Reuse the theorem number
\begin{theorem}
Suppose that Assumption \ref{assumption:Assumption B} holds with point $m^*$. Further $\eta_{1} \geq m^*_{1}$ and $\left|\eta_{2} \right| \leq m^*_{2}$. Then for spherical SGD on the given loss with $N = \alpha * d$ steps where $\alpha = \omega(1)$, $\alpha \delta^2 = o(1)$, we have that the sequence of initializations $X_0^{(d)} = \hat{v}_d$, the PCA estimators of $v_d$ obtained with $N' = \alpha'd$ unlabeled samples where $\alpha' = \omega(1)$, are Effective.
\end{theorem}
\endgroup

\begin{proof}
    The proof of Theorem~\ref{thm:theorem1} is an immediate consequence of the sequence of PCA initializations being Effective for population flow as given by Proposition \ref{prop:2} and the convergence between SGD and population flow as given by Lemma \ref{lemma:2} (below). 
\end{proof}

\begingroup
\begin{lemma}\label{lemma:2}
    Suppose that Assumption \ref{assumption:Assumption B} holds with point $m^*$. Fix any point $(m_{1}^{init}, m_{2}^{init})$ with $m_{1}^{init} > m_{1}^*, m_{2}^{init} < m_{2}^*$. For a sequence of initializations $(m_{1}(X_{0}^{d}), m_{2}(X_{0}^{d}))_{d\geq 1}$ converging to $(m_{1}^{init}, m_{2}^{init})$, spherical SGD on the given loss with the given initializations and $N = \alpha  d$ steps where $\alpha = \omega(1)$, $\alpha \delta^2 = o(1)$ yields the following:
\begin{equation}
    \sup_{t \leq N} \| (m_{1}(X_{t}), m_{2}(X_{t})) - (m_{1}(\bar{X_{t}}), m_{2}(\bar{X_{t}})) \|_{2} \to 0 \nonumber
\end{equation}

in probability as $d \to \infty$. Here $\bar{X_{t}}$ is population flow with the same initialization.
\end{lemma}
\endgroup

We will show that the spherical projections are negligible (arbitrarily small with probability 1 - o(1)). We can then consider the linearized paths of $(m_{1}(X_{t}), m_{2}(X_{t}))$ and $(m_{1}(\bar{X_{t}}), m_{2}(\bar{X_{t}}))$. Bounding their difference with Gronwall's inequality \cite{gross1967gronwall} and Doob's Inequality \cite{williams1991probability} to control the martingale term.

\begin{proof}
    Let $\nabla_E$ denote the usual Euclidean gradient. Then consider the spherical gradient: 
    
    \begin{equation*}
        \nabla \Phi(X) = \nabla_E \Phi(X) - (\nabla_E \Phi(X)\cdot X)X
    \end{equation*}
    
That is the euclidean gradient projected onto the orthogonal space of $X$. We know that the population loss can be written as a function of $(m_{1}(X)$, $m_{2}(X))$ = $(x_{1}, x_{2})$ (recall that $(m_{1}(X)$, $m_{2}(X))$ = $(x_{1}, x_{2})$ due to the without loss of generality assumption that $v_{0} = e_{1}$ and $(\eta_{2})^{-1}(v - \eta_{1}v_{0})=e_{2}$). Hence:

 \begin{equation}
      \sup_{X \in \mathbb{S}^{d-1}}\|\nabla \Phi(X)\|_{2} = \sup_{X \in \mathbb{S}^{d-1}}\|\nabla_E \Phi(X) - (\frac{\partial \phi(x_{1}, x_{2})}{\partial x_{1}}x_{1} + \frac{\partial \phi(x_{1}, x_{2})}{\partial x_{2}}x_{2})X\|_{2} \leq A \label{eq:1}
 \end{equation}

 where A is some constant independent of $d$, using that $\Phi \in C^{1}$ and $X \in \mathbb{S}^{d-1}$ (compact). Additionally with the above, one can show that there exists $K$ (independent of $d$) such that for $X,Y \in \mathbb{S}^{d-1}$ 
 
\begin{equation}
    \|\nabla\Phi(X) - \nabla\Phi(Y)\|_2 \leq K \|X - Y\|_2 \label{eq:2}
\end{equation}

We will use this fact later. Now let 
\begin{align}
    r_{t+1} & = \| X_{t} - \frac{\delta}{d}(\nabla \Phi(X_{t}) - \nabla H(X_{t}, y_t))\|_{2} \nonumber\\
    & \leq \sqrt{1 + \frac{\delta}{d}(\|\nabla \Phi(X_{t})\|_{2}^2 + \|\nabla H(X_{t}, y_t)\|_2^2)} \label{eq:3}\\
    & \leq 1 + \delta^2(\frac{A}{d^2} + \frac{L_{t}}{d})\label{eq:4}
\end{align}
 In \eqref{eq:3} we use that $X_{t}$ and $\frac{\delta}{d}(\nabla \Phi(X_{t}) - \nabla H(X_{t}, y_t))$ are orthogonal as the gradient is spherical. In \eqref{eq:4} we use that for $u>0, \sqrt{1 + u} < 1 + u$ and the spherical gradient has bounded norm shown in \eqref{eq:1}. $L_{t} = \|\nabla H(X_{t}, y_t)/ \sqrt{d}\|^2_2 $, noting that while $L_{t}$ is random, it's expectation is bounded by a constant independent of $d$ by Lemma \ref{lemma:assumption_B}. Note that the quantity we have defined $r_{t+1}$ is simply the radius of $X_{t}$ after the gradient update, but before projecting back onto the sphere. We have that $|r_{t+1} - 1| \leq \delta^2(\frac{A}{d^2} + \frac{L_{t}}{d})$. This bounds the distance between $X_t$ with itself, had it not been projected onto the sphere after the last gradient update:

\begin{equation}
    \|X_{t} - (X_{t-1} - \frac{\delta}{d}\nabla \Phi(X_{t-1}) + \frac{\delta}{d}\nabla H(X_{t-1}, y_{t-1}))\|_{2} = \frac{\left|r_{t}-1\right|}{r_t}  \|X_{t}\|_2 \leq \delta^2(\frac{A}{d^2} + \frac{L_{t}}{d}) \nonumber
\end{equation}

By iterating this bound, we have the following:

\begin{equation}
    \sup_{t \leq N} \| X_{t} - (X_{0} - \frac{\delta}{d}\sum _{i = 0}^{t-1} \nabla \Phi(X_{i}) + \frac{\delta}{d}\sum _{i = 0}^{t-1} \nabla H(X_{i}, y_i) ) \|_{2} \leq \sum _{i=0}^{N-1} \delta^2(\frac{A}{d^2} + \frac{L_{i}}{d}) \nonumber
\end{equation}

By Markov's inequality, the probability that the right hand side is greater than some $\epsilon > 0$ is:

\begin{equation}
    \mathbb{P}(\sum _{i=0}^{N-1} \delta^2(\frac{A}{d^2} + \frac{L_{i}}{d}) > \epsilon) \leq \epsilon^{-1} \alpha \delta^2 (\frac{A}{d} + \sup_{t \leq N} \mathbb{E} L_{t}) \nonumber
\end{equation}

Which is $o(1)$ given $\alpha \delta^2 = o(1)$ and $\sup_{t \leq N} \mathbb{E} L_{t} <\infty$ by Lemma \ref{lemma:assumption_B}. It thus suffices to consider the linearization of $(X_{t})_{t=0}^N$, which for the two correlation variables of interest, we denote:

\begin{equation}
    Y_{t} = \vec{m}( X_{0} - \frac{\delta}{d}\sum_{i = 0}^{t-1} (\nabla \Phi(X_{i}) - \nabla H(X_{i}, y_i)) )\nonumber
\end{equation}

where $\vec{m}(x) = (m_1(x), m_2(x))$. We note that this linearization $(Y_{t})_{t=0}^N$ is \textbf{not} the same as linear SGD. This process is equivalent to performing spherical SGD, but adding back all of the projection vectors at each stage, that were used to map $X_{t}$ to the sphere after each gradient update. Which is also \textbf{not} equivalent to doing regular gradient descent with spherical gradients. Redoing the above computations with respect to $\bar{X_{t}}$ one would see that deterministically, for large enough $d$ we have that $\bar{X}_{t}$ is also within $\epsilon$ of $\bar{Y_{t}}$, the linearization of gradient flow, given by:

\begin{equation}
    \bar{Y}_{t} = \vec{m}(\bar{X}_{0} - \frac{\delta}{d}\sum_{i = 0}^{t-1} \nabla \Phi(\bar{X}_{i})) \nonumber
\end{equation}

Hence to prove our result it is enough to show the convergence in probability between $Y_{t}$ and $\bar{Y}_{t}$. To do so, let us consider the martingale term given by $\frac{\delta}{d}\sum_{i=0}^t \nabla H(X_{t}, y_t)$. Applying Doob's inequality with $p=2$ we see that 
\begin{equation}
    \mathbb{P}(\sup_{t \leq N}\frac{\delta}{d}\|\sum_{i=0}^t \vec{m}(\nabla H(X_{t}, y_t))\|_{2} > \epsilon ) \leq \frac{2C\alpha\delta^2}{\epsilon^2 d}  = o(1/d) \nonumber
\end{equation}
 for some constant C, independent of the dimension. To show that $\sup_{t \leq N}\|Y_{t} - \bar{Y}_{t}\|_{2} \to 0$, we first consider for some fixed $T$, the quantity: $\sup_{t \leq T \delta^{-1}d}\|Y_{t} - \bar{Y}_{t}\|_{2}$. Then on the set $\left \{   \|X_{t} - Y_{t}\|_{2} \vee \|\bar{X}_{t} - \bar{Y}_{t}\|_{2} < \epsilon  \right \}$, for all $t \leq T\delta^{-1}d$:
\begin{align*}
    \|Y_{t} - \bar{Y}_{t}\|_2 &  \leq \frac{\delta}{d}\sum_{i=0}^{t-1} \|\vec{m}(\nabla \Phi(X_{i}) - \nabla \Phi(\bar{X}_{i}))\|_{2} + \frac{\delta}{d}\sum_{i=0}^{t-1} \|\vec{m}(\nabla H(X_{i}, y_i))\|_2 \\  & \leq  \frac{\delta}{d}\sum_{i=0}^{t-1} K \|\vec{m}(X_{i} - \bar{X}_{i})\|_2 + \frac{\delta}{d}\sum_{i=0}^{t-1} \| \vec{m}(\nabla H(X_{i}, y_i))\|_2 \\ & \leq 2TK\epsilon + \frac{\delta}{d}\sum_{i=0}^{t-1} K \|Y_{i} - \bar{Y}_{i}\|_2 + \frac{\delta}{d}\sum_{i=0}^{t-1} \| \vec{m}(\nabla H(X_{i}, y_i))\|_2 \\ & \leq (2TK + 1)\epsilon + \frac{\delta}{d}\sum_{i=0}^{t-1} K \|Y_{i} - \bar{Y}_{i}\|_2 
\end{align*}
with probability $1 - o(1)$, by applying Doob's inequality and noting the use of \eqref{eq:2}. Thus applying the discrete Gronwall inequality \cite{gross1967gronwall}, we obtain: 

\begin{equation}
    \sup_{t \leq T \delta^{-1}d} \|Y_{t} - \bar{Y}_{t}\|_{2} \leq (2TK + 1)\epsilon e^{KT} \label{eq:5}
\end{equation}

For any $\gamma > 0$ the above can be made less than $\gamma / 5$ by choice of $\epsilon(\gamma, T)$. We now let $T$ be such that 

\begin{equation}
    \sup_{T\delta^{-1}d \leq t \leq N} \| (m_{1}(\bar{Y}_{t}), m_{2}(\bar{Y}_{t})) - (1, 0)\|_{2} < \gamma / 5 \label{eq:6}
\end{equation}

This $T$ exists as a constant (which is independent of $d$) as a result of Assumption \ref{assumption:Assumption B} and the fact that $(m_{1}(\bar{Y}_{0}^{d}), m_{2}(\bar{Y}_{0}^{d}) )_{d\geq 1} \to (m_{1}^{init}, m_{2}^{init})$. To better understand this, recall that with constant initialization, the 2-d population dynamics of $m_{1}$ and $m_{2}$ are otherwise unaffected (other than through stepsize and number of steps) by the dimension. By Assumption \ref{assumption:Assumption B}, we have that the population dynamics converge to the intended solution at $(m_{1}, m_{2}) = (1,0)$ as $d \to \infty$. Now consider:
\begin{align}
     \sup_{ T\delta^{-1}d \leq t \leq N} \|Y_{t} - (1,0) \|_{2} 
     &\leq \|Y_{T\delta^{-1}d} - \frac{\delta}{d}\sum_{i = T\delta^{-1}d}^{t-1}\vec{m}(\nabla\Phi(X_{i})) - (1,0)\|_2 \nonumber
     + \|\frac{\delta}{d} \sum_{i=T\delta^{-1}d}^{t-1} \vec{m}(\nabla H(X_{i}, y_i))\|_2 \nonumber\\ &
     \leq 2\gamma/5 + \|\frac{\delta}{d} \sum_{i=T\delta^{-1}d}^{t-1} \vec{m}(\nabla H(X_{i}, y_i))\|_2 \nonumber
\end{align}
The above bound comes from applying triangle inequality to separate out the martingale term and then noting that at time $T\delta^{-1}d$ we have $Y_t$ is within $\gamma/5$ of $\bar{Y}_{t}$ which is within $\gamma/5$ of $(1,0)$ and the gradient with respect to the population loss only moves $Y_t$ closer to $(1,0)$. 
\begin{align}
    & \sup_{ T\delta^{-1}d \leq t \leq N} \|Y_{t} -  \bar{Y}_{t}\|_{2} \nonumber\\
    \leq & \sup_{ T\delta^{-1}d \leq t \leq N}  (\|Y_{t} - (1,0)\|_{2}  + \|\bar{Y}_{t} - (1,0)\|_{2}) \nonumber\\ \leq& 2\gamma/5 + \|\frac{\delta}{d} \sum_{i=T\delta^{-1}d}^{t-1} \vec{m}(\nabla H(X_{i}, y_i))\|_2 + \gamma/5 \leq \gamma \nonumber
\end{align}
with probability $1 - o(1)$. This follows from another application of Doob's inequality onto the projection of the high dimensional martingale onto the two fixed directions of interest $e_1, e_2$ keeping in mind Lemma \ref{lemma:assumption_B}. We have thus shown that for any $\gamma > 0$, with probability $1 - o(1)$, that $\sup_{t \leq N} \| (m_{1}(X_{t}), m_{2}(X_{t})) - (m_{1}(\bar{X_{t}}), m_{2}(\bar{X_{t}})) \|_{2} < \gamma$ which concludes the proof.
\end{proof}

\begin{proposition}\label{prop:2}
    Suppose Assumption \ref{assumption:Assumption B} holds with point $m^*$. Further $\eta_{1} \geq m^*_{1}$ and $\eta_{2} \leq m^*_{2}$. The the sequence of PCA estimators $(\hat{v}_{d})_{d \geq 1}$ of $v_d$ each obtained with $N_d = \alpha * d$ samples,$\alpha = \omega(1)$, is Effective for population flow. 
\end{proposition}

\begin{proof}
    We will use well-known facts about PCA in high dimensions, which we recall for the reader's convenance in section \ref{appendix:b} of the appendix. If we consider using a fixed linear portion of our samples for conducting PCA, we can choose the fraction $\gamma = d/N$ to be any constant we desire. Choosing $\gamma$ sufficiently small, we can ensure both that $\gamma < \lambda^2$ and $m_{1}(\hat{v}) > m^*_{1}$ and $m_{2}(\hat{v}) < m^*_{2}$. To see this consider the following:
    Let $\triangle_{1} = \left|m^*_{1} - \eta_{1}\right|$, $\triangle_{2} = \left|m^*_{2} - \eta_{2}\right|, \epsilon = \frac{1}{8}\min(\triangle_1^2, \triangle_2)^2$. We have:
  \begin{align*}
    \|\hat{v} - v\|^2_2 = \|v\|^2_2 +  \|\hat{v}\|^2_2 - 2v\cdot \hat{v}   \leq 2(1 - \frac{1 - \gamma/\lambda^2}{1 + \gamma/\lambda^2}) + \epsilon < \frac{1}{4} min(\triangle_1 ^ 2, \triangle_2^2)
  \end{align*}
for small enough $\gamma$. The second equality follows for large $d$ using a well known result regarding the limiting correlation of $v$ and $\hat{v}$ as can be seen in \citet{MDS}.

Thus by triangle inequality, we have that $m_1(\hat{v}) \geq m^*_1$ and $|m_2(\hat{v})| \leq m^*_2$ for sufficiently large $d$. Letting the choice of $\gamma$ tend to 0, it is clear the limit of $(m_1(X_0^{(d)}), m_2(X_0^{(d)}))$ exists and is equal to $(m_1(v), m_2(v)) = (\eta_1, \eta_2)$.
\end{proof}

\subsection{Proof of Theorem~\ref{thm:theorem2}}

\begingroup
\renewcommand{\thetheorem}{\ref{thm:theorem2}}
\begin{theorem}
Suppose that $f$ satisfies the following:
$$\mathbb{E}f''(g) = \mathbb{E}f'(g) = 0,\;\; \mathbb{E}\frac{\partial ^2}{\partial g^2}f(g)^2 > 0$$
for $g \sim \mathcal{N}(0,1)$. Then for spherical SGD with $N = \alpha d$ steps where $\alpha \ll d$, $\alpha \delta^2 = O(1)$, for the sequence of initializations $X_0^{(d)} \sim \mathrm{Uniform}(\mathbb{S}^{d-1})$ we have that:
$| m_1(X_N)| \to 0$
in probability, as $d \to \infty$.
\end{theorem}
\endgroup

    The overall strategy of the proof is to show that for for any $\gamma > 0$, if $\hat{X}_t = (m_1(X_t), m_2(X_t))$ is in the $d^{-1/6}$-ball, i.e. $\{ x \in \mathbb{R}^2 : \|x\|_2 <  d^{-1/6}\}$, we have that $\hat{X}_t$ enters the $\frac{1}{2}d^{-1/6}$-ball or 'times out' at $t = N$, before it leaves the $\gamma$-ball with probability $1-o(1)$.

    We want to analyze the population dynamics and we will consider doing so via a second order Taylor expansion of the population loss 
    $$\phi(x_1, x_2) = \mathbb{E}[f(a_1 x_1 + a_2 x_2 + \sqrt{1 - x_1^2 - x_2^2}g) - f(a_1)]^2 + \mathbb{E}\epsilon^2 $$ 
    in $(m_{1}(X), m_{2}(X)) = (x_1, x_2)$, around the origin. Noting that, under our assumptions on $f$, we can differentiate under the expectation, we perform the following computations:

    \begin{align*}
        & \frac{\partial \phi(0, 0)}{\partial x_1} =  \mathbb{E}2[f(g)f'(g)a_{1} - f(a_{1})f'(g)a_{1}] \\
        & \frac{\partial \phi(0, 0)}{\partial x_2} = \mathbb{E}2[f(g)f'(g)a_{2} - f(a_{1})f'(g)a_{2}] \\
        & \frac{\partial^2 \phi(0, 0)}{\partial x_1^2} = \mathbb{E}2[(f'(g)^2 +f''(g)f(g))a_{1}^2 - f(g)f'(g)g - f(a_{1})f''(g)a_{1}^2 + f(a_{1})f'(g)g] \\
        & \frac{\partial^2 \phi(0, 0)}{\partial x_2^2} = \mathbb{E}2[(f'(g)^2 +f''(g)f(g))a_{2}^2 - f(g)f'(g)g - f(a_{1})f''(g)a_{2}^2 + f(a_{1})f'(g)g]\\
        & \frac{\partial^2 \phi(0, 0)}{\partial x_1 \partial x_2} = \mathbb{E}2[(f'(g)^2 + f''(g)f(g))a_{1}a_{2} - f(a_{1})f''(g)a_{1}a_{2}]
    \end{align*}

    Evaluating the expectations we have:
    
    \begin{equation}
    \frac{\partial }{\partial x_{1}}\phi(0,0) = \mathbb{E} 2[f(g)f'(g)a_{1} - f(a_{1})f'(g)a_{1}] = -2(1 + \lambda \eta_{1}^2)\mathbb{E}f'(a_{1})\mathbb{E}f'(g) \nonumber
\end{equation}

Using that $a_{1} \perp g$, $\mathbb{E}a_{1} = 0$, and Gaussian integration by parts: $\mathbb{E}f(a_{1})a_{1} = var(a_{1})\mathbb{E}f'(a_{1})$ for $a_{1}$ Gaussian

\begin{equation}
    \frac{\partial }{\partial x_{2}}\phi(0,0) = \mathbb{E} 2[f(g)f'(g)a_{2} - f(a_{1})f'(g)a_{2}] = -2\lambda \eta_{1} \eta_{2} \mathbb{E}f'(a_{1})\mathbb{E}f'(g) \nonumber    
\end{equation}

Using that $\mathbb{E}f(a_{1})a_{2} = \mathbb{E}[f(a_{1})\mathbb{E}a_{2}|a_{1}] = \frac{\lambda \eta_{1} \eta_{2}}{1 + \lambda \eta_{1}^2} \mathbb{E}f(a_{1})a_{1}$. Which follows by another application of Gaussian integration by parts and given the following decomposition of $a_2$:
    \begin{equation}
        a_{2} = \frac{\lambda \eta_{1} \eta_{2}}{1 + \lambda \eta_{1}^2}a_{1} + \sqrt{1 + \lambda \eta_{2}^2 - \frac{(\lambda \eta_{1} \eta_{2})^2}{1 + \lambda \eta_{1}^2}} a_{1}^{\perp} \nonumber
    \end{equation}
     where $a_{1}^{\perp} \perp a_{1}$ and $var(a_{1}^{\perp})=1$.
    
 \begin{align}
  \frac{\partial^2 }{\partial x_{1}^2} \phi(0,0) &= \mathbb{E}2[(f'(g)^2 +f''(g)f(g))a_{1}^2 - f(g)f'(g)g - f(a_{1})f''(g)a_{1}^2 + f(a_{1})f'(g)g] \label{eq:7}\\
   & = 2[(1 + \lambda \eta_{1}^2)(\mathbb{E}f'(g)^2 + f''(g)f(g)) - \mathbb{E}gf(g)f'(g) - \mathbb{E}f''(g)\mathbb{E}(f(a_{1})(a_{1}^2 - 1)] \label{eq:8}\\
   & = 2[ (1+\lambda \eta_{1}^2)\mathbb{E}gf(g)f'(g) - \mathbb{E}gf(g)f'(g) - \mathbb{E}f''(g)\mathbb{E}(f(a_{1})(a_{1}^2 - 1) ] \label{eq:9}\\
   & = 2[\lambda \eta_{1}^2\mathbb{E}gf(g)f'(g) - \mathbb{E}f(g)(g^2-1)\mathbb{E}(f(a_{1})(a_{1}^2 - 1)] \nonumber
\end{align}

From \eqref{eq:7} to \eqref{eq:8} above, we used the fact that $\mathbb{E}f'(g)^2 + f''(g)f(g) = \mathbb{E}gf(g)f'(g)$ which simply follows from Gaussian integration by parts. From \eqref{eq:8} to \eqref{eq:9} we used the fact that $\mathbb{E}f''(g) = \mathbb{E}f(g)(g^2 - 1)$ which follows from two applications of Gaussian integration by parts. 

\begin{align}
    \frac{\partial^2 }{\partial x_{2}^2} \phi(0,0) & = \mathbb{E}2[(f'(g)^2 +f''(g)f(g))a_{2}^2 - f(g)f'(g)g - f(a_{1})f''(g)a_{2}^2 + f(a_{1})f'(g)g] \nonumber\\ 
    & = 2[\lambda \eta_{2}^2 \mathbb{E}gf(g)f'(g) - \mathbb{E}f''(g)\mathbb{E}f(a_{1})(a_{2}^2 - 1)] \nonumber\\
    \frac{\partial^2 }{\partial x_{1} \partial x_{2}}\phi(0,0) & = \mathbb{E}2[(f'(g)^2 + f''(g)f(g))a_{1}a_{2} - f(a_{1})f''(g)a_{1}a_{2}] \nonumber\\
     & = 2\lambda \eta_{1} \eta_{2}[\mathbb{E}gf(g)f'(g) - \mathbb{E}f''(g)\mathbb{E}f(a_{1})a_{1}^2(1 + \lambda \eta_{1}^2)^{-1} ]\nonumber
 \end{align}

Using that $\mathbb{E}f(a_{1})a_{1}a_{2} = \frac{\lambda \eta_{1} \eta_{2}}{1 + \lambda \eta_{1}^2}\mathbb{E}f(a_{1})a_{1}^2$ by tower property and decomposition of $a_2$. We use the above calculations to compute the second order Taylor expansion of $\phi$ around $(0,0)$:
\begin{align}
    \phi(x_{1}, x_{2}) & = \phi(0,0) + \frac{\partial }{\partial x_{1}}\phi(0,0)x_{1} + \frac{\partial }{\partial x_{2}}\phi(0,0)x_{2} \nonumber\\ & + \frac{1}{2}(\frac{\partial^2 }{\partial x_{1}^2}\phi(0,0)x_{1}^2 + \frac{\partial^2 }{\partial x_{2}^2}\phi(0,0)x_{2}^2 + 2\frac{\partial^2 }{\partial x_{1}\partial x_{2}}\phi(0,0)x_{1}x_{2}) + R(x_{1}, x_{2}) \nonumber
\end{align}

where $R(x_1, x_2) = O(\|(x_1, x_2)\|_2^3)$. Which allows us to compute the derivatives of $\phi(x_{1}, x_{2})$ as follows:
\begin{align*}
    \frac{\partial \phi(x_{1}, x_{2})}{\partial x_{1}} & =  \frac{\partial }{\partial x_{1}}\phi(0,0) + \frac{\partial^2 }{\partial x_{1}^2}\phi(0,0)x_{1} + \frac{\partial^2 }{\partial x_{1} \partial x_{2}}\phi(0,0)x_{2} + O(\|(x_1, x_2)\|_2^2) \\
    & = -2(1 + \lambda \eta_{1}^2)\mathbb{E}f'(a_{1})\mathbb{E}f'(g) \nonumber\\ & + 2[\lambda \eta_{1}^2\mathbb{E}gf(g)f'(g) - \mathbb{E}f(g)(g^2-1)\mathbb{E}(f(a_{1})(a_{1}^2 - 1)]x_{1} \nonumber\\ & + 2\lambda \eta_{1} \eta_{2}[\mathbb{E}gf(g)f'(g) - \mathbb{E}f''(g)\mathbb{E}f(a_{1})a_{1}^2(1 + \lambda \eta_{1}^2)^{-1}]x_{2} + O(\|(x_1, x_2)\|_2^2)
\end{align*}

\begin{align}
    \frac{\partial \phi(x_{1}, x_{2})}{\partial x_{2}} & =  \frac{\partial }{\partial x_{2}}\phi(0,0) + \frac{\partial^2 }{\partial x_{2}^2}\phi(0,0)x_{2} + \frac{\partial^2 }{\partial x_{1} \partial x_{2}}\phi(0,0)x_{1} + O(\|(x_1, x_2)\|_2^2) \nonumber\\ & = -2\lambda \eta_{1} \eta_{2} \mathbb{E}f'(a_{1})\mathbb{E}f'(g) \nonumber\\ & + 2[\lambda \eta_{2}^2 \mathbb{E}gf(g)f'(g) - \mathbb{E}f''(g)\mathbb{E}f(a_{1})(a_{2}^2 - 1)]x_{2} \nonumber\\ & + 2\lambda \eta_{1} \eta_{2}[\mathbb{E}gf(g)f'(g) - \mathbb{E}f''(g)\mathbb{E}f(a_{1})a_{1}^2(1 + \lambda \eta_{1}^2)^{-1}]x_{1} \nonumber + O(\|(x_1, x_2)\|_2^2) \nonumber
\end{align}

Using our assumptions on $\mathbb{E}f''(g), \mathbb{E}f'(g)$, we have the simplified system:

\begin{equation}
    \frac{\partial \phi(x_{1}, x_{2})}{\partial x_{1}} =  2\lambda \eta_{1}^2\mathbb{E}gf(g)f'(g)x_{1} + 2\lambda \eta_{1} \eta_{2}\mathbb{E}gf(g)f'(g)x_{2} + O(\|(x_1, x_2)\|_2^2) \label{eq:10}
\end{equation}
    
\begin{equation}
    \frac{\partial \phi(x_{1}, x_{2})}{\partial x_{2}} =  2\lambda \eta_{2}^2 \mathbb{E}gf(g)f'(g)x_{2} + 2\lambda \eta_{1} \eta_{2}\mathbb{E}gf(g)f'(g)x_{1} + O(\|(x_1, x_2)\|_2^2) \label{eq:11}
\end{equation}

We recall that the spherical gradient correction term is higher order:

\begin{equation*}
(\frac{\partial \phi(x_{1}, x_{2})}{\partial x_{1}}x_{1} + \frac{\partial \phi(x_{1}, x_{2})}{\partial x_{2}}x_{2})X 
\end{equation*}

 Analyzing the linearized system, given by \eqref{eq:10} and \eqref{eq:11}, we see that the first order terms are orthogonal to the line $L = \{ tv_2^\perp: t \in \mathbb{R}\}=\{ (x_1,x_2): x_2 = -\frac{\eta_1}{\eta_2} x_1\}$, where $v_2^\perp = (\eta_2, -\eta_1)$. On the line $L$, the first order terms are both 0, and hence the magnitude of the first order effects tends to 0 as $(x_1, x_2)$ tends to the line $L$. The magnitude of the first order terms, exceeds the magnitude of all higher order terms when the distance of $(x_1, x_1)$ to the line $L$, exceeds $c\|(x_1,x_2)\|_2^2$ for some constant $c$. 

Before proceeding we define some notation. Let us introduce the four quadrants $Q_1 = \{(x,y): x > 0, y > 0\},Q_2 = \{(x,y): x < 0, y > 0\},Q_3 = \{(x,y): x < 0, y < 0\}, Q_4 = \{(x,y): x > 0, y < 0\}$. We will consider the variable $\hat{X}_t = \vec{m}(X_t) = (m_1(X_t), m_2(X_t)) = (x_1, x_2)_t$. 

We define two operators $T_{L1}$ and $T_{L2}$ as follows: $T_{L1}(\hat{X}) = v_2^\perp \cdot P_L$ where $P_L$ is the orthogonal projection operator onto $L$. $T_{L1}$ is defined such that for $\hat{X}$ in $Q_4$, $T_{L1}(\hat{X}) > 0$. We define $T_{L2} = - T_{L1}$ so that $T_{L2}(\hat{X}) > 0$ for $\hat{X} \in Q_2$. We also define the 'left and right half-spaces' with respect to $L$ as follows: $H_l = \{(x_1, x_2): x_2 < - \frac{\eta_1}{\eta_2}x_1\}$ and $H_r = \{(x_1, x_2): x_2 > - \frac{\eta_1}{\eta_2}x_1\}$. 

We now define two additional operators $T_{r}^\perp$ and $T_{l}^\perp$ as follows. Let $v_2 = (\eta_1, \eta_2)$ and then let $T_{r}^\perp(\hat{X}) = v_2 \cdot (I - P_L)(\hat{X})$. $T_{r}^\perp$ measures the signed distance of $\hat{X}$ to the line $L$, with $T_{r}^\perp(\hat{X}) > 0$ when $\hat{X} \in Q_1$. Note that $T_{L1}, T_{L2}, T_{r}^\perp, T_l^{\perp}$ are all linear operators.

With our notation above we now define the following set: $\mathcal{C} = \{(x_1, x_2): |T_{l}^\perp(x_1, x_2)| < c \|(x_1, x_2)\|_2^2\}$. For $\hat{X} \notin C$, we have that the first order terms in the gradient of the population loss, exceed the higher order terms.

Let $L_{1}^*$ be the line $\{ (x_1,x_2): x_2 = -\eta_1/2\eta_2 x_1\}$ if $\eta_1 > \eta_2 $ or $\{ (x_1,x_2): x_2 = -2\eta_1/\eta_2 x_1\}$ if $\eta_1 < \eta_2 $. 
In essence this line is the line half way between $L$ and whichever quadrant boundary is closer to $L$.
We define it's counter part $L_2^*$ as the line with the reciprocal slope, i.e. the slope of $L_2^*$ is 1 over the slope of $L_1^*$. 
Without loss of generality we assume $\eta_1 < \eta_2$. Similar to the 'left and right half-spaces' with respect to $L$ defined above, we can define the left and right half-spaces of $L_1^*$ and $L_2^*$, $H_{l1}, H_{r1}, H_{l2}, H_{r2}$. 
Then we consider the set $Q^* = (H_{l2} \cap H_{r1}) \cup (H_{r2} \cap H_{l1})$. We now notice that if we consider the set $\mathcal{C}$ intersect the $\gamma$-ball, that for sufficiently small $\gamma$, the intersection is a subset of $Q^*$. For a better understanding of the definitions given above see Figure \ref{appendix_guide2}. We will make use of the set $\mathcal{C}$ in Lemma \ref{negative_lemma_2} below.

We now state three lemmas which we will use to complete our proof of Theorem \ref{thm:theorem2}. We defer the proofs of these lemmas until after the proof of Theorem \ref{thm:theorem2}. Lemma \ref{negative_lemma_3} will tell us that when $\hat{X}_t$ is in $Q_1$ or $Q_3$, it must leave, entering $Q_2$ or $Q_4$ before it's norm grows by $d^{-1/3}$. Lemma \ref{negative_lemma_1} will tell us that $\hat{X}$ cannot leave the $\gamma$-ball through $Q_2$ or $Q_3$ without first exiting the quadrant, provided it arrived small enough. Lemma \ref{thm:theorem2} provides a bound on how far $\hat{X}$ can move away from the line $L$ before entering the set $\mathcal{C}$ defined above. Together these lemmas allow us to control the magnitude of $\hat{X}_t$ before it enters the set $\mathcal{C}$, and then show that once $\hat{X}_t$ is in $\mathcal{C}$ it cannot escape the $\gamma$-ball before re-entering the $1/2d^{-1/6}$-ball.

\begin{figure}[H] 
\begin{center}
\centerline{\includegraphics[width=\columnwidth]{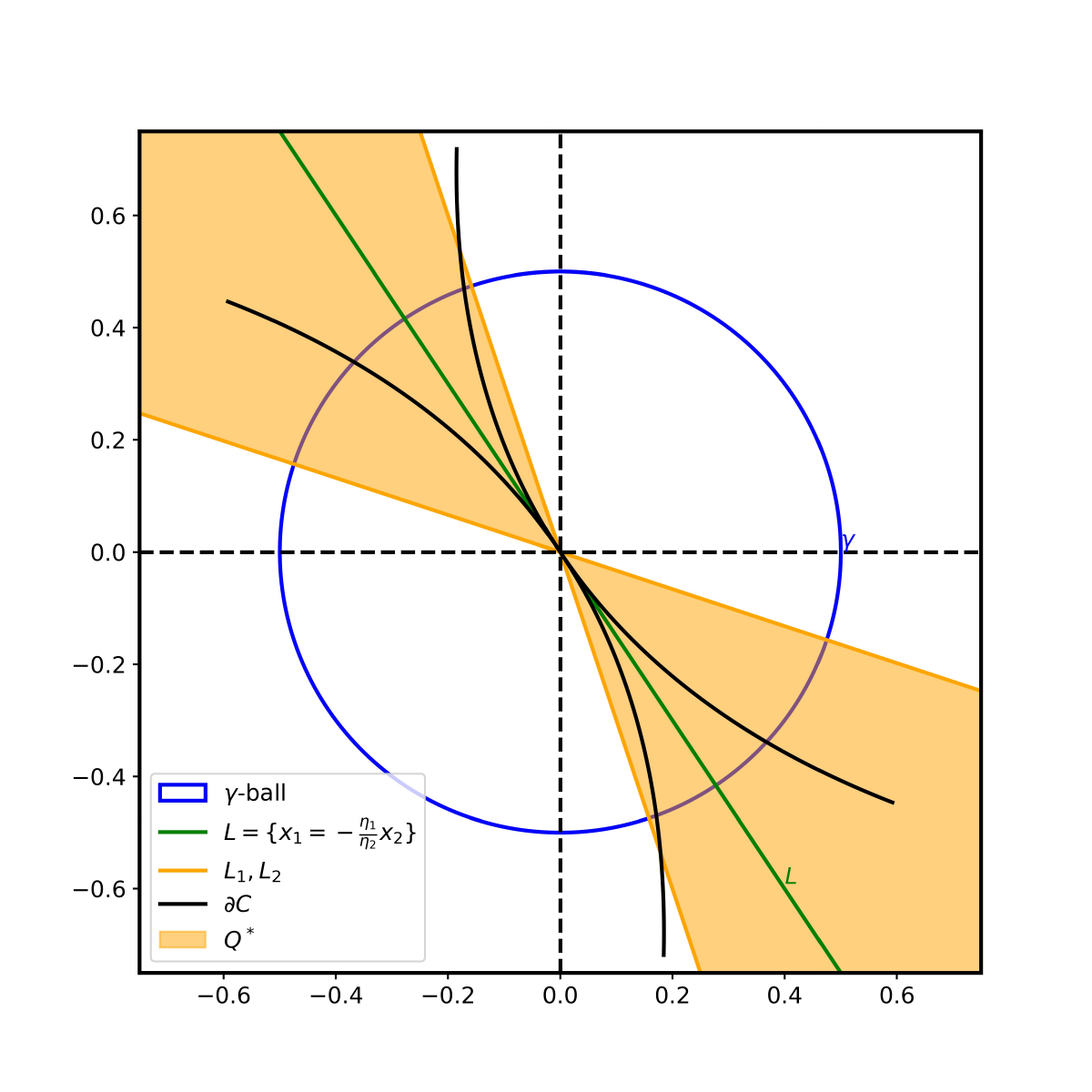}}
\caption{A visual to help guide in understanding the definitions of $L_1, L_2$ and the set $\mathcal{C}$. It is clear that regardless of $\eta_1$, by decreasing the vale of $\gamma$, we find that $\mathcal{C}$ intersect the $\gamma$-ball is a subset of $Q^*$}
\label{appendix_guide2}
\end{center}
\end{figure}
    
\begin{figure}[H] 
\begin{center}
\centerline{\includegraphics[width=\columnwidth]{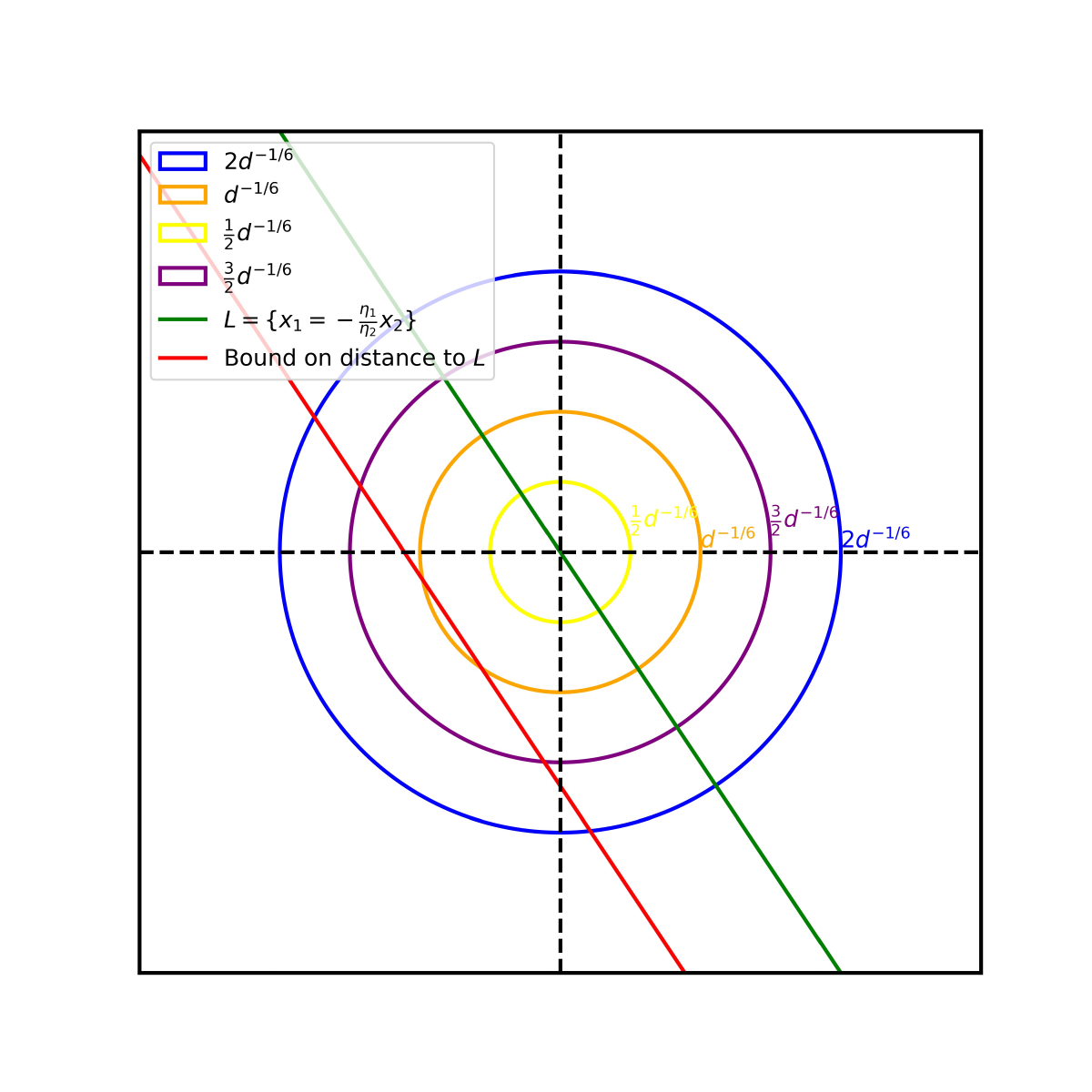}}
\caption{A visual to help guide in the understanding of the proof of Theorem \ref{thm:theorem2} and the use of the three lemmas. When initialized in $Q_3$ in the $d^{-1/6}$-ball, $\hat{X}$ must leave $Q_3$ before leaving the $3/2d^{-1/6}$-ball. Suppose it enters $Q_4$, the red line provides a boundary $\hat{X_t}$ cannot cross before entering the set $\mathcal{C}$ (see Figure \ref{appendix_guide2}).}
\label{appendix_guide}
\end{center}
\end{figure}

\begin{lemma}\label{negative_lemma_1}
     There exists some sequence $K_d$ growing to infinity such that under the assumptions of Theorem \ref{thm:theorem2} and restricting to the set $E_1 = \{\sup_{t \leq N} |\frac{\delta}{d}\sum_{i=1}^{t}\nabla T_{L1}\vec{m}((H(X_i)))| < \frac{1}{2}K_d/\sqrt{d} \}$, we have that if  $\hat{X}_{t^*} \in Q_4,$ and $\|\hat{X}_{t^*}\|_2 \leq d^{-1/10}$, then 
    \begin{equation}
        \|\hat{X_t}\|_2 < \gamma, \forall t \in [t^*, \min(\tau_{Q_4^-}, N)]
    \end{equation}
         where $\tau_{Q_4^-}$ is the stopping time for the next time $\hat{X_t}$ leaves $Q_4$. On that event, the same statement is true with $Q_2$ in place of $Q_4$ as well. Further, the $\mathbb{P}(E_1) = 1-o(1)$. 
\end{lemma}

This lemma says that under the set $E_1$, for $\hat{X_t}$ with norm less than $d^{-1/10}$ and in $Q_4$, $\hat{X_t}$ must leave the quadrant $Q_4$ before it leaves the $\gamma$-ball. Similarly, for $\hat{X}$ in $Q_2$ with norm less than $d^{-1/10}$, $\hat{X}$ exits $Q_2$ before it leaves the $\gamma$-ball.

\begin{lemma}\label{negative_lemma_2}
Letting $\hat{X}^\perp = T_{l}^\perp (\hat{X})$,
 under the assumptions of Theorem \ref{thm:theorem2} under the event: $E_2 = \{\sup_{t \leq N} \|\frac{\delta}{d}\sum_{i=1}^{t}\nabla T_{l}^{\perp}(\vec{m}(H(X_i)))\|_2 < \frac{1}{2}d^{-1/3} \}$, when $\hat{X}_{t^*} \in H_l, \;\| \hat{X}_{t^*}\|_2 \leq 2d^{-1/6}$ we have that:

        \begin{equation*}
          \|\hat{X}_t^\perp \|_2 \leq \|\hat{X}_{t^*}^\perp \|_2 + d^{-1/3}, \forall t \in [t^*, \min(\tau_{C}, \tau_{1/2d^{-1/6}},N)] 
        \end{equation*}

        where $\tau_C$ is the next time $\hat{X}_t$ enters the set $\mathcal{C}$ and $\tau_{1/2d^{-1/6}}$ is the next time $\hat{X}_t$ enters the $1/2 d^{-1/6}$-ball. Further, the same statement holds when replacing $T_{l}^\perp$ with $T_{r}^\perp$ and $H_l$ with $H_r$.        
\end{lemma}

    This lemma says that if $\hat{X}_t$ is in the $2d^{-1/6}$ ball, then it's distance from the line $L$ can only increase by up to $d^{-1/3}$ before it enters the set $\mathcal{C}$ or re-enters the $1/2 d^{-1/6}$-ball.

\begin{lemma}\label{negative_lemma_3}
Under the assumptions of Theorem \ref{thm:theorem2} and on the sets $E^*_{j} = \{\sup_{t \leq N} |\frac{\delta}{d}\sum_{i=1}^{t}\nabla H(X_i) \cdot e_j| < \frac{1}{10}d^{-2/5} \}, \; j = 1,2$, when $\hat{X}_{t^*} \in Q_3$, we have that:

    \begin{equation*}
      \|\hat{X}_t\|_2 \leq \|\hat{X}_{t^*}\|_2 + d^{-1/5}, \forall t \in [t^*, \tau_{Q_3^-}] 
    \end{equation*}

Where $\tau_{Q_3^-}$ is the stopping time for the next time $\hat{X}_t$ leaves the quadrant $Q_3$. Further the above statement holds replacing $Q_3$ with $Q_1$.
\end{lemma}

    This lemma says that the maximum amount the norm of $\hat{X}_t$ can increase while in $Q_3$ before leaving is $d^{-1/5}$.

\begin{proof}[Proof of Theorem~\ref{thm:theorem2}]
  We intend to show that for any $\gamma > 0$,  $\mathbb{P}(\sup_{t\leq N} \|(m_1(X_t), m_2(X_t)))\|_2 > \gamma) \to 0$ as $d \to \infty$.

    The proof of this result will make use of the three lemmas above to show that if $\hat{X}_t$ is in the $d^{-1/6}$-ball, it will re-enter the $d1/2^{-1/6}$ ball before it leaves the $\gamma$ ball. To make use of the three lemmas above, we note that the sets considered in these lemmas: $E_1, E_2, E^*_1, E^*_2$ are all probability $1-o(1)$ by Doob's inequality and hence so is their intersection $E = E_1 \cap E_2 \cap E^*_1 \cap E^*_2$. We now remind the reader that random initializations yield correlations on the order of $m_1(X_0), m_2(X_0) = O(\frac{1}{\sqrt{d}}) \ll d^{-1/6}$. We now argue that under the set $E$ for $\hat{X}_t$ in the $d^{-1/6}-ball$, we have deterministically that $\hat{X}_t$ re-enters the $1/2 d^{-1/6}$ ball or reaches $t = N$ before it leaves the $\gamma$-ball, hence by the Markov property, this will conclude the proof.
    
    We will prove this in a case by case manner, firstly considering the case that $\hat{X}_t$ is in the $d^{-1/6}$-ball and in $Q_3$. 
    
    By Lemma \ref{negative_lemma_3}, $\hat{X}_t$ must leave $Q_3$ before it can leave the $3/2d^{-1/6}$-ball. Hence for $\hat{X}_t$ to leave the $\gamma$-ball, it must first exit the quadrant. So suppose $\hat{X}_t$ enters $Q_4$ (the case where $\hat{X}_t$ enters $Q_2$ follows by a similar argument). Since it entered small enough (by small enough we mean $\|\hat{X}_t \|_2 \leq d^{-1/10}$, allowing us to invoke lemma \ref{negative_lemma_1}), by Lemma \ref{negative_lemma_1}, it cannot leave the $\gamma$-ball before exiting the quadrant.  
    Additionally, by Lemma \ref{negative_lemma_2}, it's distance from $L$ cannot increase by more than $d^{-1/3}$ before entering the set $\mathcal{C}$.
    Now also notice that for $\hat{X}_t$ to enter $Q_3 \setminus  2d^{-1/6}$-ball, it would require it's distance from $L$ to increase by a quantity that is order $d^{-1/6} \gg d^{-1/3}$ which cannot happen before $\hat{X}_t$ enters $\mathcal{C}$. 
    Thus $\hat{X}_t$ cannot exit the $\gamma$-ball through $Q_3$ until entering $\mathcal{C}$. $\hat{X}_t$ may re-enter $Q_3$ or even $Q_2$, but any entry to $Q_4$ or $Q_2$ requires $\|\hat{X}_t\|_2 \leq d^{-1/10}$ and hence by by lemma \ref{negative_lemma_1}, $\hat{X}_t$ cannot exit the $\gamma$-ball through quadrants $Q_2$ or $Q_4$. Hence $\hat{X}_t$ cannot exit the $\gamma$-ball through any quadrant without re-entering the $1/2 d^{-1/6}$-ball, timing out or first entering the set $\mathcal{C}$. So suppose $\hat{X}_t$ enters $\mathcal{C}$ in $Q_4$. $\hat{X}_t$ still cannot leave the $\gamma-ball$ until first leaving the quadrant. But now recall that $\mathcal{C}$ is contained in $Q^*$ and hence to leave the quadrant without first re-entering the $1/2d^{-1/6}$-ball, requires $\hat{X}_t$ to exit $Q^*$, and further move a distance on the order of $d^{-1/6}$ away from the line, which by lemma \ref{negative_lemma_2} cannot happen. This completes the proof for the case where $\hat{X}_t$ is in $Q_3$. Note that the case where $\hat{X}_t$ is in $Q_1$ follows by symmetric arguments. Now finally notice that the cases where $\hat{X}_t$ is in either $Q_2$ or $Q_4$ follow by the same arguments as above, noting that the $Q_1$ and $Q_3$ cases reduce to the $Q_2$ and $Q_4$ cases.
    
    \end{proof}

% \begingroup
% \renewcommand{\thetheorem}{\ref{negative_lemma_1}}
% \begin{lemma}
%      There exists some sequence $K_d$ growing to infinity such that under the assumptions of Theorem \ref{thm:theorem2} and restricting to the set $E_1 = \{\sup_{t \leq N} |\frac{\delta}{d}\sum_{i=1}^{t}\nabla T_{L1}\vec{m}((H(X_i)))| < \frac{1}{2}K_d/\sqrt{d} \}$, we have that if  $\hat{X}_{t^*} \in Q_4,$ and $\|\hat{X}_{t^*}\|_2 \leq d^{-1/10}$, then 
%     \begin{equation}
%         \|\hat{X_t}\|_2 < \gamma, \forall t \in [t^*, \min(\tau_{Q_4^-}, N)]
%     \end{equation}
%          where $\tau_{Q_4^-}$ is the stopping time for the next time $\hat{X_t}$ leaves $Q_4$. On that event, the same statement is true with $Q_2$ in place of $Q_4$ as well. Further, the $\mathbb{P}(E_1) = 1-o(1)$. 
% \end{lemma}
% \endgroup

% This lemma says that under the set $E_1$, for $\hat{X_t}$ with norm less than $d^{-1/10}$ and in $Q_4$, $\hat{X_t}$ must leave the quadrant $Q_4$ before it leaves the $\gamma$-ball. Similarly, for $\hat{X}$ in $Q_2$ with norm less than $d^{-1/10}$, $\hat{X}$ exits $Q_2$ before it leaves the $\gamma$-ball.

\begin{proof}[Proof of Lemma~\ref{negative_lemma_1}]
    We will prove the case where $\hat{X} \in Q_4$ using $T_{L1}$ and simply note that the case of $\hat{X} \in Q_2$ with $T_{L2}$ follows by the same arguments.
    
     Note that for any $\eta_1$ and any value $\gamma > 0$, $\exists \epsilon_\gamma$ such that for $x \in Q_4 , \|x\|_2 > \gamma \Rightarrow T_{L1}(x) > \epsilon_\gamma$. We will now show that the event $E_1$ contains the following event:
    \begin{equation}
        \{T_{L1}\hat{X_t} < \epsilon_\gamma, \forall t \in [t^*, \min(\tau_{Q_4^-}, N)]\} \nonumber
    \end{equation}
Observe that for such $t \in [t^*, \min(\tau_{Q_4^-}, N)]$:
    \begin{align}
    T_{L1}(\hat{X}_t) & \leq T_{L1}(\hat{X}_{t-1} -  \frac{\delta}{d}\vec{m}(\nabla \Phi(X_{t-1})) - \frac{\delta}{d}\vec{m}(\nabla H(X_{t-1}))) \nonumber \\ & \leq T_{L1}(\hat{X}_{t^*}) - T_{L1}(\frac{\delta}{d}\sum_{i=t^*}^{t-1}\vec{m}(\nabla \Phi(X_{i})) + \frac{\delta}{d}\sum_{i=t^*}^{t-1}\vec{m}(\nabla H(X_{i}))) \nonumber
  \end{align}
    Note that this follows from the fact that $T_{L1}\hat{X}_t > 0$ for as long as $\hat{X}_t \in Q_4$, and that the radius of $X_t$ plus a gradient step, is deterministically greater than one as it is a spherical gradient. Now given $\hat{X}_t \in Q_4$ and recalling that the first order terms in the gradient are orthogonal to the line $L$, we have that $T_{L1}(\vec{m}(\nabla \Phi (X_t)))$ is at most second order in $T_{L1}(\hat{X}_t)$. Hence:
    \begin{equation}
        T_{L1}(\hat{X}_t) \leq T_{L1}(\hat{X}_{t^*}) + c \frac{\delta}{d}\sum_{i=t^*}^{t-1} |T_{L1}\hat{X}_i|^2 - T_{L1}(\frac{\delta}{d}\sum_{i=t^*}^{t-1}\vec{m}(\nabla H(X_{i})))) \nonumber
    \end{equation}
    For some constant $c$. Note that the event$\{ \sup_{t \leq N} |\frac{\delta}{d}\sum_{i=1}^{t}\nabla T_{L1}(\vec{m}(H(X_i)))| < \frac{1}{2}K_d/\sqrt{d}\}$ contains the event $\{ \sup_{j,t : 1 \leq i \leq t\leq N} |\frac{\delta}{d}\sum_{i=j}^{t}\nabla T_{L1}(\vec{m}(H(X_i)))| < K_d/\sqrt{d}\}$. Thus we have that:

    \begin{equation}
        T_{L1}(\hat{X}_t) \leq  T_{L1}(\hat{X}_{t^*}) + c \frac{\delta}{d}\sum_{i=t^*}^{t-1} |T_{L1}\hat{X}_i|^2 + K_d/\sqrt{d}, \forall t \in [t^*, \min(\tau_{Q_4^-}, N)] \nonumber
    \end{equation}
    
  It follows from this inequality that $T_{L1}\hat{X}_t < \epsilon_\gamma, \: \forall t \in [t^*,\min(\tau_{Q_4^-}, N)] $. To see this, observe that by the discrete Bihari-LaSalle inequality \ref{bihari_appendix} that for some constant $c_1$:

 \begin{equation}
     T_{L1}(\hat{X}_t) \leq  2 \frac{K_d}{\sqrt{d}} (1 - c_1 \delta K_d d^{-3/2}t)^{-1}      \nonumber
 \end{equation}

so long as $T_{L1}(\hat{X}_{t^*}) \leq \frac{K_d}{\sqrt{d}}$ (which is true so long as $d^{-\frac{1}{10}} \leq \frac{K_d}{\sqrt{d}}$). For sufficiently large $d$, the right hand side above is smaller than $\epsilon_\gamma$ for all $t \leq \hat{t}$ where:
\begin{equation}
    \hat{t} = K_d ^{-\epsilon}\delta^{-1} d ^{3/2} \nonumber
\end{equation}
    for some $\epsilon > 0$ sufficiently small, provided $K_d$ is growing slower than $d^{1/2 - \zeta}$ for some $\zeta > 0$. 
    
    Thus if we choose $K_d$ to be diverging appropriately slowly, we have that $\hat{t} > N$ (recalling that $\sqrt{\alpha} = o(\sqrt{d})$ and $\sqrt{\alpha} = O(\delta^{-1})$). We thus choose $K_d$ such that $T_{L1}(\hat{X}_{t^*}) \leq d^{-\frac{1}{10}} \leq \frac{K_d}{\sqrt{d}}$ and $K_d \leq d^{1/2 - \zeta}$ (for some $\zeta > 0$), for example $K_d = d^{1/2 - 1/11}$, then $  K_d/\sqrt{d} = d^{-1/11} \gg d^{-1/10}$ and further $\mathbb{P}(\sup_{t \leq N} |\frac{\delta}{d}\sum_{i=1}^{t}\nabla T_{L1}\vec{m}((H(X_i)))| < \frac{1}{2}K_d/\sqrt{d}) \geq 1 - O(\alpha \delta^2 / K_d^2) = 1 - o(1)$.
    \end{proof}

%\begingroup
%\renewcommand{\thetheorem}{\ref{negative_lemma_2}}
%\begin{lemma}
 %Letting $\hat{X}^\perp = T_{l}^\perp (\hat{X})$,
 %under the assumptions of Theorem \ref{thm:theorem2} under the event: $E_2 = \{\sup_{t \leq N} \|\frac{\delta}{d}\sum_{i=1}^{t}\nabla T_{l}^{\perp}(\vec{m}(H(X_i)))\|_2 < \frac{1}{2}d^{-1/3} \}$, when $\hat{X}_{t^*} \in H_l, \;\| \hat{X}_{t^*}\|_2 \leq 2d^{-1/6}$ we have that:

%        \begin{equation*}
%          \|\hat{X}_t^\perp \|_2 \leq \|\hat{X}_{t^*}^\perp \|_2 + d^{-1/3}, \forall t \in [t^*, \min(\tau_{C}, \tau_{1/2d^{-1/6}},N)] 
%        \end{equation*}

%        where $\tau_C$ is the next time $\hat{X}_t$ enters the set $\mathcal{C}$ and $\tau_{1/2d^{-1/6}}$ is the next time $\hat{X}_t$ enters the $1/2 d^{-1/6}$-ball. Further, the same statement holds when replacing $T_{l}^\perp$ with $T_{r}^\perp$ and $H_l$ with $H_r$.
%\end{lemma}
%\endgroup
%    This lemma says that if $\hat{X}_t$ is in the $2d^{-1/6}$ ball, then it's distance from the line $L$ can only increase by up to $d^{-1/3}$ before it enters the set $\mathcal{C}$ or re-enters the $1/2 d^{-1/6}$-ball. 

    \begin{proof} [Proof of Lemma~\ref{negative_lemma_2}]

    We prove the result in the case that $\hat{X}_{t^*} \in H_l$ where $\hat{X}_{t^*}^\perp = T_{l}^\perp (\hat{X}_{t^*})$ noting that this implies $\hat{X}_{t} \in H_l \; \forall t \in [t^*, \min(\tau_{C}, \tau_{1/2d^{-1/6}},N)]$. The case where $\hat{X}_{t^*} \in H_r$ and replacing $T_{l}^\perp$ with $T_{r}^\perp$, follows by an identical argument. Observe that
    \begin{align*}
    \hat{X}_t^\perp  \leq \hat{X}_{t-1}^\perp - T_{l}^\perp(\vec{m}(\frac{\delta}{d}\nabla \Phi(X_{t-1}))) - T_{l}^\perp(\vec{m}(\frac{\delta}{d}\nabla H(X_{t-1})))  \leq  \hat{X}_{t-2}^\perp - T_{l}^\perp(\vec{m}(\frac{\delta}{d}\nabla H(X_{t-1}) + \frac{\delta}{d}\nabla H(X_{t-2})))
        \end{align*}
    Once again the first inequality comes from the fact that the spherical gradient is always greater than 1 and $\hat{X}_t^\perp > 0$ for $\hat{X}_t$ in $H_l$. The second inequality follows from two observations. The first observation is that removing the spherical projections provides an upper bound given $\hat{X}_{t}^\perp > 0$ in the time interval considered here. The second observation is that removing the gradient of the population loss term, also provides an upper bound. This follows from our analysis of the population dynamics (provided in the discussion before the proof), which told us that the first order terms in the population dynamics, point orthogonally towards the line $L$. Hence when the first order terms exceed the higher order terms (i.e. when $\hat{X_t} \notin C$), the gradient of the population loss term, is negative under the operator $T_{l}^\perp$ when $\hat{X} \in H_l$. 

    Expanding the above we have:
    \begin{align*}
    \|\hat{X}_t^\perp\|_2  \leq \| \hat{X}_{t^*}^\perp\|_2 + \| T_{l}^\perp(\frac{\delta}{d}\sum_{i=t^*}^{t}\nabla H(X_i))\|_2   \leq  \| \hat{X}_{t^*}^\perp\|_2 + d^{-1/3} 
    \end{align*}
under $E_2$.
    \end{proof}

%\begingroup
%\renewcommand{\thetheorem}{\ref{negative_lemma_3}}
%\begin{lemma}
%Under the assumptions of Theorem \ref{thm:theorem2} and on the sets $E^*_{j} = \{\sup_{t \leq N} |\frac{\delta}{d}\sum_{i=1}^{t}\nabla H(X_i) \cdot e_j| < \frac{1}{10}d^{-2/5} \}, \; j = 1,2$, when $\hat{X}_{t^*} \in Q_3$, we have that:

%        \begin{equation*}
%          \|\hat{X}_t\|_2 \leq \|\hat{X}_{t^*}\|_2 + d^{-1/5}, \forall t \in [t^*, \tau_{Q_3^-}] 
%        \end{equation*}

%        Where $\tau_{Q_3^-}$ is the stopping time for the next time $\hat{X}_t$ leaves the quadrant $Q_3$. Further the above statement holds replacing $Q_3$ with $Q_1$.
%\end{lemma}
%\endgroup
%    This lemma says that the maximum amount the norm of $\hat{X}_t$ can increase while in $Q_3$ before leaving is $d^{-1/5}$.

    \begin{proof} [Proof of Lemma~\ref{negative_lemma_3}]
         We prove the case of $Q_1$ and make note that the case of $Q_3$ follows by a symmetric argument. We note that in $Q_1$ we have that $m_1(X_t)$, $m_2(X_t) > 0$ and $\nabla \Phi(X_t) \cdot e_1 > 0, \nabla \Phi(X_t) \cdot e_2 > 0 $ (this follows from the discussion on the population dynamics, given prior to the proof of \ref{thm:theorem2}). We now consider $m_1(X_t)= X_t \cdot e_1$:
\begin{align}
 X_t\cdot e_1 & \leq X_{t-1}\cdot e_1 - \frac{\delta}{d}\nabla \Phi(X_{t-1})\cdot e_1 - \frac{\delta}{d}\nabla H(X_{t-1})\cdot e_1 \nonumber \\ & \leq X_{t^*}\cdot e_1 - \frac{\delta}{d} \sum_{i = t^*}^{t}\nabla\Phi(X_{i})\cdot e_1 - \frac{\delta}{d} \sum_{i = t^*}^{t}\nabla H(X_{i})\cdot e_1\nonumber \\ & \leq X_{t^*}\cdot e_1 - \frac{\delta}{d}\sum_{i = t^*}^{t}\nabla H(X_{i})\cdot e_1\nonumber \\ & \leq X_{t^*}\cdot e_1 + \frac{1}{5}d^{-2/5} \nonumber
\end{align}
 The inequalities above follow by similar arguments to the previous two lemmas noting that the sign of $X_t \cdot e_1$ and the gradients are always the same while in $Q_1$. The bound on the martingale term follows under the set $E^*_1$. A similar argument follows for $X_t \cdot e_2$. We conclude the proof noting that:

 \begin{align}
     \| \hat{X}_t \|_2 & \leq \sqrt{(X_{t^*} \cdot e_1 + \frac{1}{5}d^{-2/5})^2 + (X_{t^*} \cdot e_2 + \frac{1}{5}d^{-2/5})^2} \nonumber \\ & \leq \sqrt{(X_{t^*} \cdot e_1)^2 + (X_{t^*} \cdot e_2)^2 + d^{-2/5}} \nonumber \\ & \leq \sqrt{\|\hat{X}_{t^*}\|_2^2 + d^{-2/5} + 2d^{-1/5}\|\hat{X}_{t^*}\|_2} \nonumber \\ & = \sqrt{(\|\hat{X}_{t^*}\|_2 + d^{-1/5})^2} \nonumber \\ & = \|\hat{X}_{t^*}\|_2 + d^{-1/5} \nonumber
 \end{align}
        
    \end{proof}

\subsection{Proof of Theorem~\ref{thm:theorem3}}
\begingroup
\renewcommand{\thetheorem}{\ref{thm:theorem3}}
\begin{theorem}
Suppose that $f$ satisfies the following:
$$\mathbb{E}f''(g) = \mathbb{E}f'(g) = 0,\;\; \mathbb{E}\frac{\partial ^2}{\partial g^2}f(g)^2 > 0$$
for $g \sim \mathcal{N}(0,1)$. When $\eta_1 = 1$, for spherical SGD with $N = \alpha d$ steps where $\alpha = \omega(1)$, $\alpha \delta^2 \ll d^{1/3}$ , we have that there exists some $r > 0$ such that for all sequences of initializations $X_0^{(d)}\; : \; \left| m_1(X_0^{(d)}) \right|< r \; \forall \; d$, then we have that:
\begin{equation*}    \left|m_1(X_N)\right| \to 0 \nonumber
\end{equation*}
in probability as $d \to \infty$. Further we have that for all $\epsilon > 0$:
\begin{equation*}
    \mathbb{P}(\sup_{t \leq N}\left|m_1(X_t)\right| > r + \epsilon) \to 0 
\end{equation*}
in probability as $d \to \infty$.

\end{theorem}
\endgroup

\begin{proof}
 From the Taylor expansion used in the proof of Theorem \ref{thm:theorem2}, specifically equations \eqref{eq:10} and \eqref{eq:11}, we see that for $\eta_1 = 1 \Rightarrow \eta_2 = 0$, the first order term in the population dynamics sends $x_1$ to 0. This is to say that $\exists r > 0 : \forall X \;s.t.\; m_1(X) \in \mathbb{B}(0,2r)$ (the 1-dimensional open ball of radius $2r$) we have that $\nabla \Phi(X)\cdot v_0 sgn(X\cdot v_0)>0$. This tells us that the population gradient step sends $x_1$ towards 0, for all $x_1$ in the specified ball. We note that the value of $r$ is independent of the dimension.
 
We now start by restricting to the set $E = \{\sup_{t\leq N}|\frac{\delta}{d}\sum_{i = 1}^{t}\nabla H(X_i)\cdot v_0| < d^{-1/3}\}$ . We note that by Doob's inequality we have that: $$\mathbb{P}(\sup_{t\leq N}|\frac{\delta}{d}\sum_{i = 1}^{t}\nabla H(X_i)\cdot v_0| > d^{-1/3}) \leq C \frac{\alpha\delta^2}{d^{1/3}} = o(1)$$ (recalling the assumption that $\alpha\delta^2 \ll d^{1/3}$. Then note that the set $E$ implies the set $\{\sup_{0 \leq j \leq t \leq N} |\frac{\delta}{d}\sum_{i = j}^{t}\nabla H(X_i)\cdot v_0| < 2d^{- 1/3}\}$.

We now claim that under the set $E$, we have that deterministically $m_1(X_N) \to 0$ as $d \to \infty$. Consider the sequence of stopping times, $(\tau_k)_{k\geq1}$, corresponding to where $m_1(X_t)$ crosses zero, i.e., let $\tau_k = \inf \{t > \tau_{k-1}:sgn(m_1(X_t)) \neq sgn(m_1(X_{t-1}))\}$ and $\tau_0 = 0$. Note the maximum single step that $m_1(X_t)$ can take is bounded and tending to 0 in the dimension (recall the stepsize is $\frac{\delta}{d}$ and the gradient of the loss is bounded, as seen in Lemma \ref{lemma:assumption_B} and the proof of Theorem \ref{thm:theorem1} . Thus we have that $m_1(X_{\tau_k}) \to 0$ as $d \to \infty$ whenever $\tau_k < N$. This is because the distance of $m_1(X_{\tau_k})$ to 0 is bounded by the maximum single step.

We will proceed by breaking up the interval $[0,N]$ into the union of $\cup_{k\geq 0} [\tau_k, \min(\tau_{k+1}, N+1))$. We separately consider the first interval $[0, \min(\tau_{1},N+1))$ and each other interval $[\tau_k, \min(\tau_{k+1}, N+1)),\; k\geq 0$, starting with the latter.

Fix $k>0$ and suppose that $t\in[\tau_k,\min(\tau_k+1,N+1))$. Suppose, without loss of generality, that $m_1(X_{\tau_k}) \geq 0$. Now for $t \in [\tau_k, \min(\tau_{k+1}, N+1))$, this process remains positive and we have that:

    \begin{align*}
    m_1(X_{t+1}) &= X_{t+1} \cdot v_0  \\
    &= \frac{ \left( X_t - \frac{\delta}{d} \left( \nabla \Phi(X_t) + \nabla H(X_t) \right) \right) \cdot v_0 }{\|X_t - \frac{\delta}{d} \left( \nabla \Phi(X_t) + \nabla H(X_t) \right) \|_2}  \\ 
    &\leq  \left( X_t - \frac{\delta}{d} \left( \nabla \Phi(X_t) + \nabla H(X_t) \right) \right) \cdot v_0  \\ 
    & \leq (X_{\tau_k} - \frac{\delta}{d}\sum_{i=\tau_k}^t \left(\nabla \Phi(X_i)+\nabla H(X_i) )\right)\cdot v_0  \\ 
    & \leq (X_{\tau_k} - \frac{\delta}{d}\sum_{i=\tau_k}^t \nabla \Phi(X_i))\cdot v_0 + 2d^{-1/3} \\ & \leq m_1(X_{\tau_k}) + 2d^{-1/3} \\ & \leq 4d^{-1/3}
    \end{align*}

The first inequality follows from the fact that the spherical gradient step always results in a point with norm greater than or equal to 1. 
The second inequality follows from the fact that removing the spherical projections for each $m_1(X_{t})$ provides an upper bound as the process is positive over the time interval considered. The third inequality simply applies the restricting set $E$. The fourth inequality comes from the fact that the $\nabla \Phi(X_t))\cdot v_0 > 0$ whenever $0 < m_1(X_t) < 2r$. The final inequality comes from the maximum one step change of $m_1(X_t)$ (which is $O(\frac{\delta}{d}) \ll d^{-1/3}$) which provides an upper bound on $m_1(X_{\tau_k})$ for $k>0$.

Now to see that $m_1(X_t) < 2r, \; \forall t \in [\tau_k, \min(\tau_{k+1}, N+1))$, we suppose for the sake of a contradiction, that $m_1(X_t) > 2r$ for some time $t^* \in [\tau_k, \min(\tau_{k+1}, N+1))$ and further $t^*$ is the first time $m_1(X_t)$ exceeds $2r$. Repeating the inequalities above (noting that $m_1(X_t) < 2r, \; \forall t \in [\tau_k, t^*)$) we have that \[m_1(X_{t^*}) \leq 4d^{-1/3} \leq 2r
\] a contradiction. We thus have that for any interval $[\tau_k, \min(\tau_{k+1}, N+1))$, the value of $m_1(X_t)$ is upper bounded by $4d^{-1/3}$. Note that we only showed this for the case $m_1(X_{\tau_k}) > 0$, but the case $m_1(X_{\tau_k}) < 0$ follows by a similar argument. 

The first statement of the proof, i.e. that $m_1(X_N) \to 0$ in probability as $d \to \infty$ would then follow so long as $\tau_1 < N+1$. Which is to say that if the process $m_1(X_t)$ crosses 0 at least once, it will remain within $d^{-1/3}$ of 0 and hence be there at time $N$, proving the first statement of the Theorem. The second statement of the proof would also follow so long as $\sup_{t\leq \min(\tau_1, N+1)}|m_1(X_t)| < r + \epsilon$ for any fixed $\epsilon > 0$, deterministically under the restricted set $E$, which has probability $1-o(1)$.  

We now proceed to finish the proof by considering the time interval $[0, \min(\tau_1,N+1))$. We once again assume without loss of generality that $m_1(X_0) > 0$, i.e. that the process is positive over the time interval considered. Then we again have for $t$ in this interval:
    \begin{align}
    m_1(X_{t+1}) &= X_{t+1} \cdot v_0  \nonumber\\
    &= \frac{ \left( X_t - \frac{\delta}{d} \left( \nabla \Phi(X_t) + \nabla H(X_t) \right) \right) \cdot v_0 }{\left\| \left( X_t - \frac{\delta}{d} \left( \nabla \Phi(X_t) + \nabla H(X_t) \right) \right) \cdot v_0 \right\|_2} \nonumber \\ &\leq  \left( X_t - \frac{\delta}{d} \left( \nabla \Phi(X_t) + \nabla H(X_t) \right) \right) \cdot v_0  \nonumber \\ & \leq (X_{0} - \frac{\delta}{d}\sum_{i=0}^t \nabla \Phi(X_i))\cdot v_0  +  \frac{\delta}{d}\sum_{i=0}^t \nabla H(X_i) \cdot v_0 \nonumber \\ & \leq (X_{0} - \frac{\delta}{d}\sum_{i=0}^t \nabla \Phi(X_i))\cdot v_0 + 2d^{-1/3} \nonumber
    \end{align}

    Once again, so long as $m_1(X_t) \leq 2r$, we have that $\nabla \Phi(X_i))\cdot v_0 > 0$. By the same contradiction based argument as before, we can show that $m_1(X_t) \leq 2r$ for all $t \in [0, \min(\tau_1,N+1))$ (recalling that $m_1(X_0) < r$). Hence we have the bound:
\[
 m_1(X_{t+1}) \leq r + 2d^{-1/3}
\]
 Which completes the proof of the second statement of the Theorem, i.e. that for all $\epsilon > 0$:
\begin{equation*}
    \mathbb{P}(\sup_{t \leq N}\left|m_1(X_t)\right| > r + \epsilon) \to 0 
\end{equation*}
in probability as $d \to \infty$. We now complete the proof of the first statement of the theorem by showing that $m_1(X_N) \to 0$ in probability as $d \to \infty$ in the case that $\tau_1 > N$. Suppose for the sake of contradiction that there exists some $c > 0$ such that:

\begin{equation}\label{eq:thm3.3}    
\limsup_{d \to \infty}\mathbb{P}(\inf_{t\leq N} |m_1(X_t)| > c) > 0
\end{equation}

Before proceeding, consider the converse of this assumption which is that for every $c > 0$:

\begin{align*}
 & \limsup_{d \to \infty}\mathbb{P}(\inf_{t\leq N} |m_1(X_t)| > c) = 0 \\  \Rightarrow \;\; & \mathbb{P}(\inf_{t\leq N} |m_1(X_t)| \leq c) = 1 - o(1)
\end{align*}

Then fixing any $c > 0$ and restricting to the $1-o(1)$ probability set $\{\inf_{t\leq N} |m_1(X_t)| \leq c\} \cap E$, let $\tau_c = \inf \{t: |m_1(X_t)| \leq c\}$. Assume without loss of generality that $m_1(X_t) > 0, \; t \in [0, \tau_c ]$. Once again repeating the same sequence of inequalities used previously and the same contradiction argument that allows us to invoke those inequalities, we have that:
\[
m_1(X_N) \leq m_1(X_{\tau_c}) + 2d^{-1/3} \leq c +  2d^{-1/3} \leq 2c
\]
Which is to say that for any $c>0$, $m_1(X_N) \leq 2c$ with probability $1-o(1)$, which would complete the proof. Hence, we return to the assumption given by \eqref{eq:thm3.3}. Now recall that we have already shown that restricting to $E \cap \{\tau_1<N+1\}$ we have $m_1(X_t) \leq 4d^{-1/3}$ deterministically, we thus restrict to the set $E \cap \{\tau_1 > N\}$. Now we may assume that $0 \leq m_1(X_t) \leq 2r$ for all $t \in [0,N)$. The lower bound follows under $\{\tau_1<N+1\}$, the upper bound $m_1(X_t) < 2r$ holds generally with probability $1-o(1)$ due to the second statement of the theorem with $\epsilon = r$, which has already been proven at this point. Now by the previous analysis of the population loss via Taylor expansion, we have that there exists some constant $c_1$ such that $\nabla \Phi(X_t) \cdot v_0 \geq c_1 m_1(X_t)$ for all $t \in [0,N)$ and further on the subsequence $d_k, k\geq 1$ such that: 
$$\lim_{d_k \to \infty}\mathbb{P}(\inf_{t \leq N}|m_1(X_t)| > c) > 0$$
 if we restrict to the positive probability set that $\{\inf_{t \leq N}|m_1(X_t)| > c\}$ again assuming without loss of generality $m_1(X_t) > 0 \;\forall \; t \in [0,N]$, we have $\Phi(X_t) \cdot v_0 \geq c_1c$ for all $t \in [0,N)$. We then see:
    \begin{align*}
         m_1(X_t) & \leq  (X_{0} - \frac{\delta}{d}\sum_{i=0}^t \nabla \Phi(X_i))\cdot v_0 + 2d_k^{-1/3} \\ & \leq (X_{0} - \frac{\delta}{d}\sum_{i=0}^N c_1 c) \cdot v_0 + 2d_k^{-1/3}\\ & \leq m_1(X_0) - cc_1\alpha\delta + 2d_k^{-1/3}
    \end{align*}
    Which is diverging to $-\infty$ as $d_k \to \infty$. This is a contradiction as this places a negative upper bound on $m_1(X_t)$ which is strictly positive on the positive probability set considered. This completes the proof.

\end{proof}

\subsection{Proof of Lemma \ref{lemma:hermite} and Theorem~\ref{thm:TL}}

\begin{lemma}\label{lemma:hermite}
    For all Hermite Polynomials with degree 3 or greater, $h_k(x), k \geq 3$: $$\mathbb{E}h_k''(g) = \mathbb{E}h_k'(g) = 0,\;\; \mathbb{E}\frac{\partial ^2}{\partial g^2}f(g)^2 > 0$$
for $g \sim \mathcal{N}(0,1)$
\end{lemma}

\begin{proof}
    For all Hermite polynomials with degree 1 or greater, $\mathbb{E}h_k = 0$. Further it is well known the Hermite polynomials satisfy the following:
    
    \begin{align*}
        &h_n^{(k)}(x) = \frac{n!}{(n-k)!}h_{n-k}(x) \\ \Rightarrow &\mathbb{E}h_n^{(k)}(g) = \frac{k!}{(n-k)!}\mathbb{E}h_{n-k}(g) = 0, \forall k<n
    \end{align*}

    This gives us that $\mathbb{E}h_k''(g) = \mathbb{E}h_k'(g) = 0$. Now consider $\mathbb{E}\frac{\partial ^2}{\partial g^2}f(g)^2$ = $\mathbb{E}h_k'(g)^2 + \mathbb{E}h_k''(g)h_k(g)$. Now we have that:
    $$\mathbb{E}h_k''(g)h_k(g) = k(k-1)\mathbb{E}h_k(g)h_{k-1}(g) = 0$$ by orthogonality. This completes the claim.
    
\end{proof}

%\begingroup
%\renewcommand{thetheorem}{\ref{thm:TL}}
% \begin{theorem}\label{thm:TL}
%     Suppose Assumption $f$ is differentiable almost everywhere with $f'$ at most polynomial growth. Let $k \geq 2$ be the information exponent of $\Phi$. Let $v^{(d)} \cdot v_{0}^{(d)} = \eta_d = \theta(d^{-\zeta})$, with $\zeta \in [0,1/2)$. Then for spherical SGD with $N=\alpha d$ steps with $\alpha \gg d^{2\zeta(k-2)} log(d)^{2 \textbf{1}[\zeta > 0]}$ and $\alpha^{-1} \ll \delta \ll \alpha^{-1/2} $, $X_0 = v$, we have that: $$\left|m_1(X) - 1\right| \to 0$$
%     in probability as $d \to \infty$. Here $\textbf{1}[\zeta > 0]$ is the indicator function, taking value 0 if $\zeta = 0$ and 1 otherwise.

% \end{theorem}
%\endgroup

\begin{proof}[Proof of Theorem~\ref{thm:TL}]
    The proof of this theorem follows from \cite{arous2021online}. Specifically in section 2.1 of their paper they show that the single-index model covered here meets the assumptions required for there main results which apply more generally.
    Then noting that by replacing a random uniform initialization of order $d^{-1/2}$ with a fixed initialization of order $d^{-\zeta}$, all their arguments for Theorem 1.3 of their paper still hold, with the new sample complexity provided above (depending on $\zeta$). In the case of $\zeta = 0$, the result simply follows by applying Theorem 3.2 of their paper, noting the arguments used to prove this theorem apply just as well when considering a sequence of initializations which itself is not constant but is bounded above and below by constants.
\end{proof}

\section{PCA in High Dimensions}
\label{appendix:b}
We consider applying PCA in high dimensions in Theorem \ref{thm:theorem1}. There are a number of well known results about applying PCA in the high dimensional limit including the BBP transition\cite{baik2005phase}. We informally state a few of these results here for completeness, however, we refer the reader to \citet{MDS} for more details. 

Consider a $d$-dimensional isotropic Gaussian vector $Z \sim N(0, I_{d})$. Letting $X \in \mathbb{R}^{d \times n}$ denote the data matrix ($n$ rows of observations of $Z$). If we let both $n$ and $d$ grow to infinity, keeping their ratio fixed $d/n = \gamma$, the distribution of the eigenvalues of the matrix $\frac{1}{n}X^TX$ (the sample covariance) will in the limit, follow the Marcenko-Pastur distribution given by: $$ dF_{\gamma}(x) = \frac{1}{2\pi \gamma x}\sqrt{(\gamma_{+} - x)(x - \gamma_{-})(\gamma x)}\mathbf{1}[\gamma_{-},\gamma_{+}](x)dx $$ with $\gamma_{+} = (1+\gamma)^2$ and $\gamma_{-} = (1 - \gamma)^2$. Thus in high dimensions, we can expect to observe top eigenvalues of $\frac{1}{n}X^TX$ up to size $(1+\frac{d}{n})^2$, even when there is no covariance structure on $X$ at all. In order to detect the spike we require $\lambda \geq \sqrt{d/n}$. The limiting squared correlation between the top eigenvector of the sample covariance matrix and the true spike can be shown to be $\left|v \cdot \hat{v}\right|^2 = \frac{1 - \gamma / \lambda^2}{1 + \gamma / \lambda^2}$ when $\lambda \geq \sqrt{d/n}$ and 0 otherwise.

\section{Discrete Bihari-LaSalle Inequality}\label{bihari_appendix}

The discrete Bihari-LaSalle Inequality claims that for a sequence $m_t$ satisfying the following for some $k \geq 2, a,b >0$:
\[
m_t \leq a + \sum_{i=0}^{t-1}bm_i^{k-1}
\]
then we have that $m_t \geq \frac{a}{(1 - ba^{k-2}t)^{\frac{1}{k-2}}}$

For a proof see Appendix C of \cite{arous2021online}.

%\section{Additional Figure}
%\begin{figure}[!t]
%\begin{tabular}{cc}
 % \includegraphics[width=140mm]{paths_3.pdf} \\  \includegraphics[width=130mm]{paths_4.pdf}
%\end{tabular}

%\caption{The figures above display the results of the second experiment and correspond to $\eta_1 = 1, \lambda = 0.5$. The paths of four independent SGD runs are observed. One random initialization, one initialization via pre-training with PCA and 2 fixed initializations. The fixed initializations are $m_1(X_0) = 0.1$ and $m_1(X_0) = 0.25$. In addition to noticing the behavior of random initialization versus pre-training, we observe very different behavior for fixed initializations. For small enough $m_1(X_0)$ SGD tends towards the local optima at $m_1(X) = 0$ as suggested by Theorem \ref{thm:theorem3}. For larger enough $m_1(X_0)$, SGD tends towards the global optima.}
%\label{appendix_simulations}
%\end{figure}

\section{Effect of $\eta$ on Learning}

To investigate the impact of the strength in correlation between the spike and parameter vectors, $\eta_1$, we run the following experiment. We fix a function $f$ which we again take as the third Hermite polynomial, and we vary the value of $\eta_1$ and perform training with PCA initialization (and all other settings the same as described in section 5.2. We observe as expected that for large enough values of $\eta_1$ SGD with PCA initialization is able to recover the unknown parameter vector and there is a point at which $\eta_1$ becomes too small and SGD is unable to recover even with PCA initialization. See the figure below.

\begin{figure}[t]
\begin{center}
\includegraphics[width=130mm]{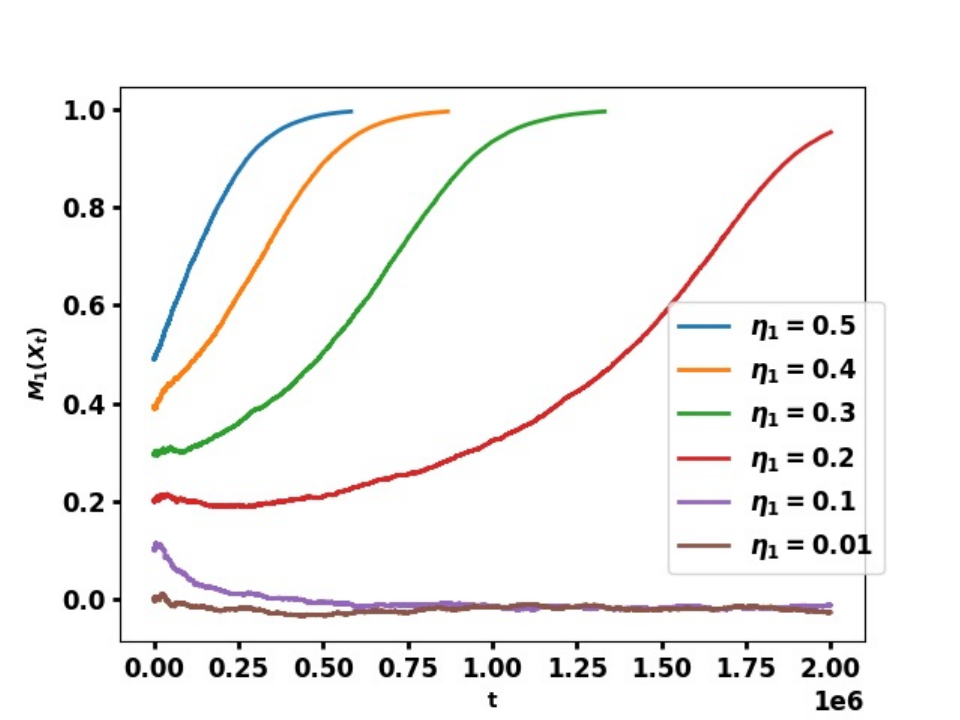}

\caption{The figure above displays the correlation values of $M_1(X_t)$ over the course of training via SGD with PCA initializations. As expected we see that there is some threshold for $\eta_1$, below which the spike vector provides insufficient information about the final parameter vector at initialization for SGD to realize the true parameter vector on a reasonable time scale. }
\end{center}
\end{figure}

\end{document}